%% file: main.tex
\documentclass[10pt,journal,compsoc]{IEEEtran}
\usepackage[pdftex]{graphicx}
\usepackage{pdfpages}
\usepackage{amsmath}
\usepackage{multirow}

\ifCLASSOPTIONcompsoc
  \usepackage[nocompress]{cite}
\else
  \usepackage{cite}
\fi
\begin{document}
\title{Sketch Beautification: Learning Part Beautification and Structure
Refinement for Sketches of Man-made Objects}

\author{Deng~Yu,
        Manfred~Lau\IEEEauthorrefmark{1},
        Lin~Gao,
        and~Hongbo~Fu\thanks{\IEEEauthorrefmark{1} Corresponding authors}\IEEEauthorrefmark{1}
        % ,~\IEEEmembership{Member,~IEEE}% <-this % stops a space
\IEEEcompsocitemizethanks{
\IEEEcompsocthanksitem Deng Yu, Manfred Lau, and Hongbo Fu are with the School
of Creative Media, City University of Hong Kong.\protect\\
E-mail: \{deng.yu@my, manfred.lau@, hongbofu@\}cityu.edu.hk
\IEEEcompsocthanksitem  Lin Gao is with the Beijing Key Laboratory of Mobile Computing and Pervasive Device, Institute of Computing Technology, Chinese Academy of Sciences, Beijing, China, and also with the University of Chinese Academy of Sciences, Beijing, China.\\
E-mail: gaolin@ict.ac.cn}
% <-this % stops an unwanted space
\thanks{Manuscript received xx xx, xx; revised xx xx, xx.}}


\markboth{IEEE TRANSACTIONS ON VISUALIZATION AND COMPUTER GRAPHICS}%
{Shell \MakeLowercase{\textit{et al.}}: Bare Demo of IEEEtran.cls for Computer Society Journals}



\IEEEtitleabstractindextext{%
% \begin{abstract}
% The abstract goes here.
% \end{abstract}
\input{sections/abstract.tex}
}

\maketitle

\IEEEdisplaynontitleabstractindextext

% \fi
\IEEEpeerreviewmaketitle

\IEEEraisesectionheading{
% \section{Introduction}\label{sec:introduction}
% \cite{bae2008ilovesketch}.
}
\input{sections/introduction.tex}
\input{sections/related_work.tex}
\input{sections/methodology.tex}

\input{sections/experiments.tex}

\input{sections/conclusion.tex}
\input{sections/limitation_and_discussion.tex}

\ifCLASSOPTIONcaptionsoff
  \newpage
\fi



\bibliographystyle{IEEEtran}
\bibliography{bibpapers}
% \input{bibpapers.bib}
% \input{bibpapers.bib}
  
% \end{thebibliography}

\vfill
% \begin{IEEEbiography}{Deng Yu}
\begin{IEEEbiography}[{\includegraphics[width=1in,height=1.25in,clip,keepaspectratio]{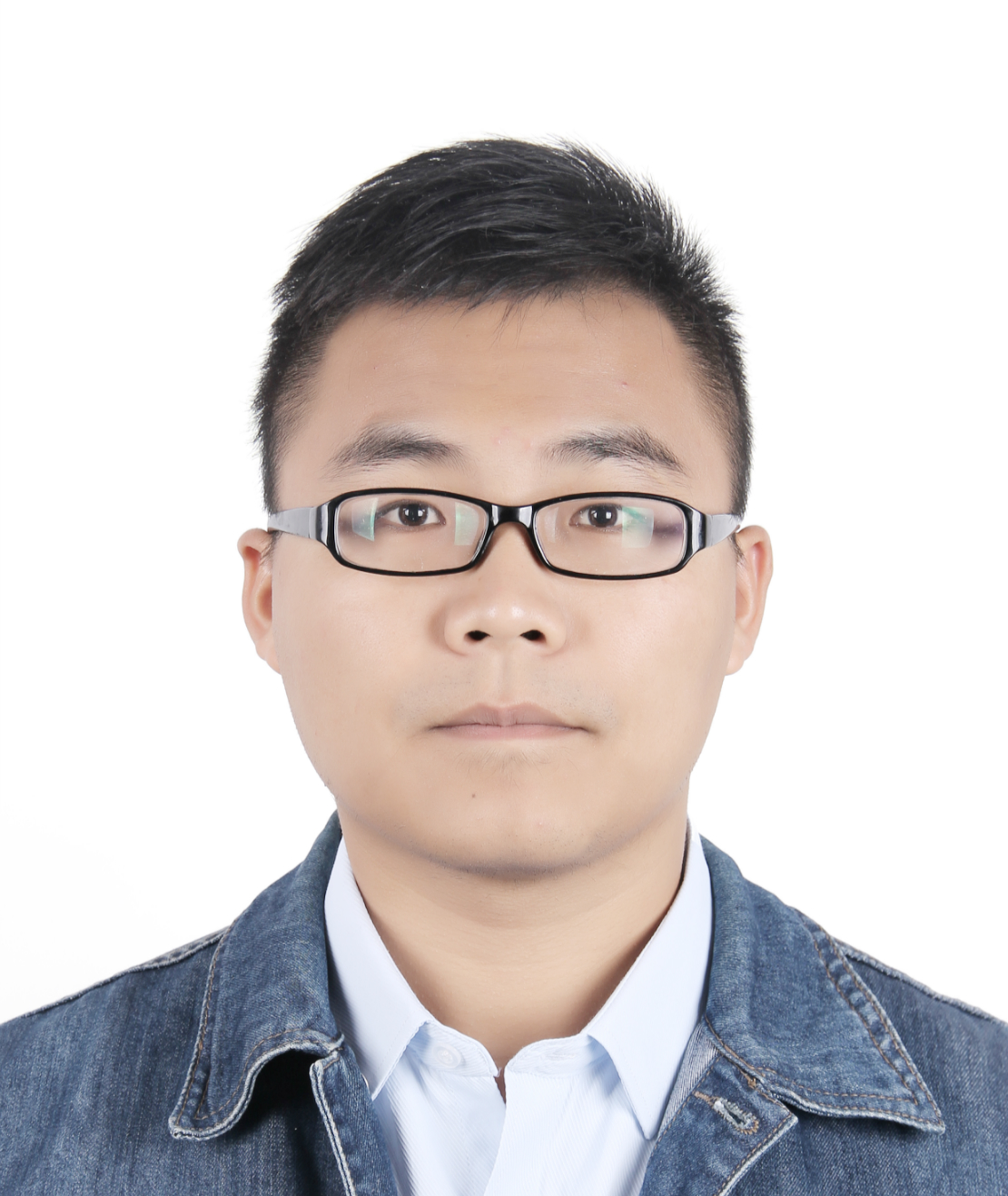}}]{Deng Yu}
is pursuing the Ph.D. degree at
the School of Creative Media, City University of Hong Kong. He received a B.Eng. degree and a Master's degree in computer science and technology from China University of Petroleum (East China). His research interests include computer graphics and data-driven techniques.
\end{IEEEbiography}
\vfill
\begin{IEEEbiography}[{\includegraphics[width=1in,height=1.25in,clip,keepaspectratio]{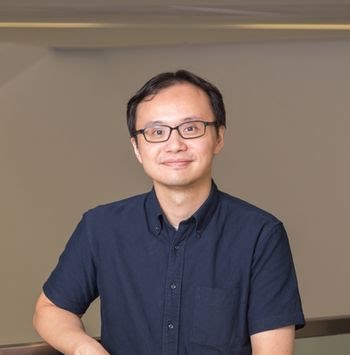}}]{Manfred Lau}
is an Assistant Professor in the School of Creative Media at the City University of Hong Kong. His research interests are in computer graphics, human-computer interaction, and digital fabrication. His recent research in the perception of 3D shapes uses crowdsourcing and learning methods for studying human perceptual notions of 3D shapes. He was previously an Assistant Professor in the School of Computing and Communications at Lancaster University in the UK, and a post-doc researcher in Tokyo at the Japan Science and Technology Agency - Igarashi Design Interface Project. He received his Ph.D. degree in Computer Science from Carnegie Mellon University, and his B.Sc. degree in Computer Science from Yale University. He has served on the program committees of major graphics conferences including Siggraph Asia.
\end{IEEEbiography}
\vfill
%\newpage
\begin{IEEEbiography}[{\includegraphics[width=1in,height=1.25in,clip,keepaspectratio]{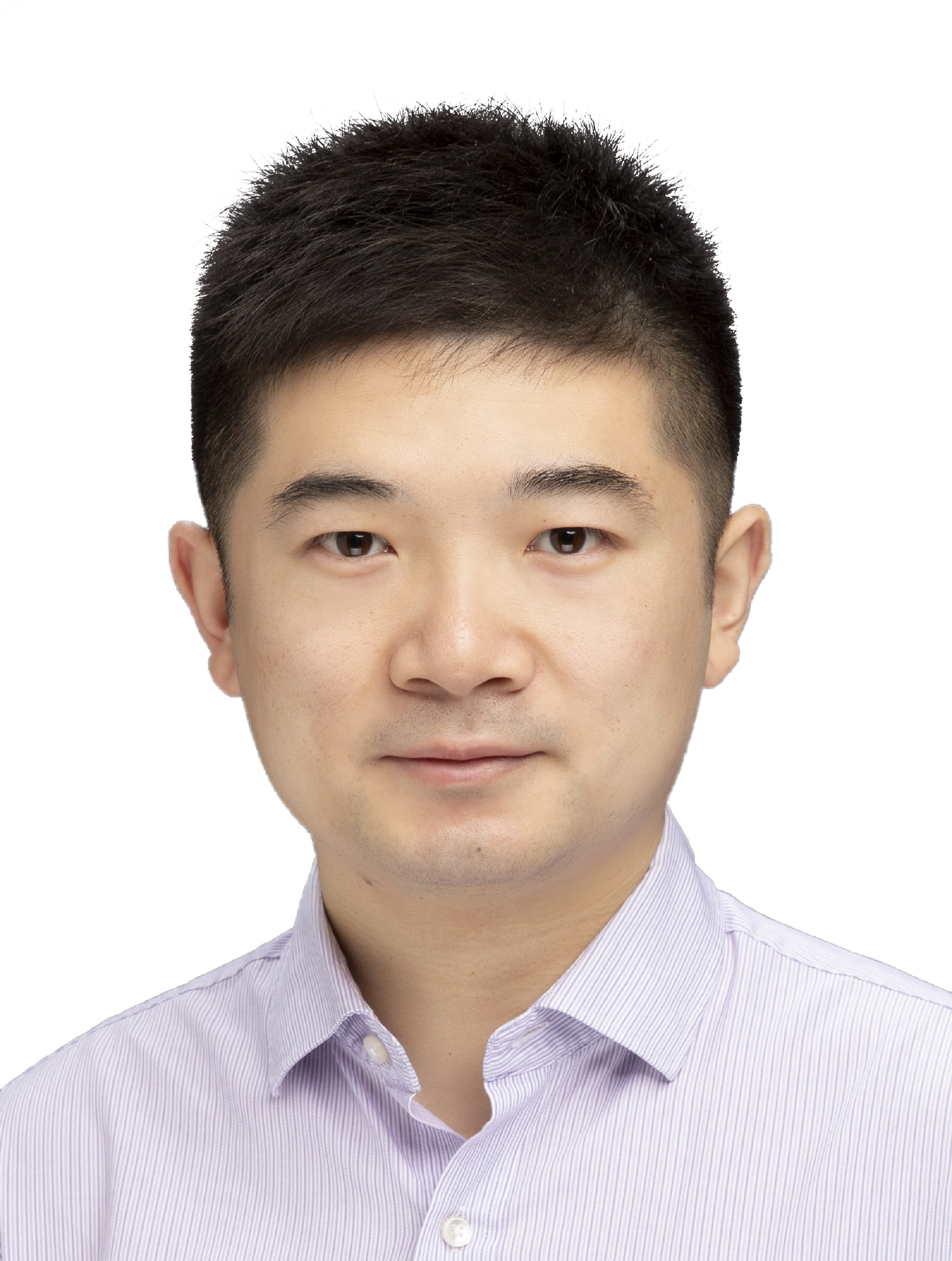}}]{Lin Gao} received his PhD degree in computer science from Tsinghua University. He is currently a Professor at the Institute of Computing Technology, Chinese Academy of Sciences. He has been awarded the Newton Advanced Fellowship from the Royal Society and the AG Young Researcher Award. His research interests include computer graphics and geometric processing.

\end{IEEEbiography}
\vfill
\begin{IEEEbiography}[{\includegraphics[width=1in,height=1.25in,clip,keepaspectratio]{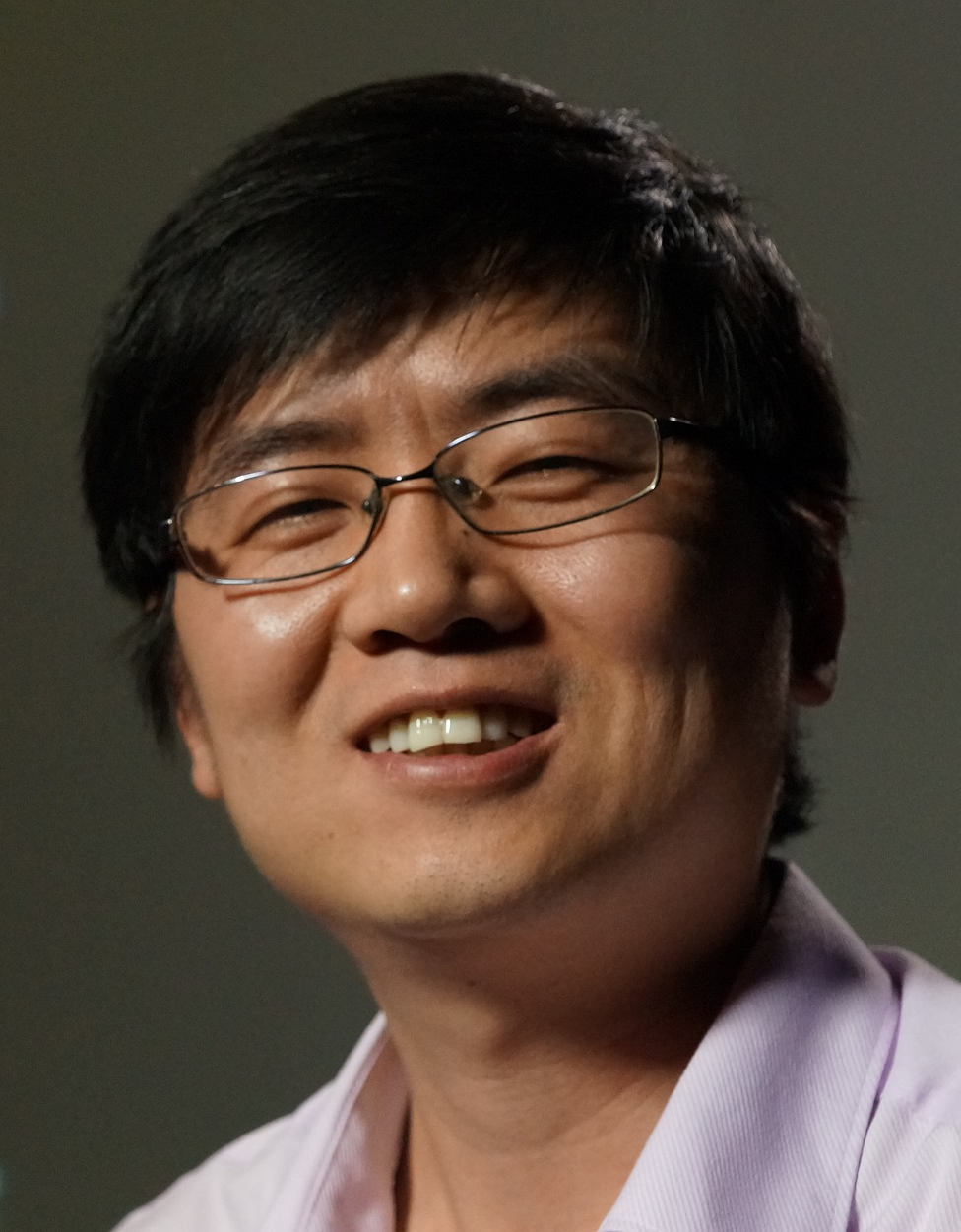}}]{Hongbo Fu}
is a Professor at the School of
Creative Media, City University of Hong Kong.
He received a B.S. degree in information sciences from Peking University and a Ph.D.
degree in computer science from Hong Kong
University of Science and Technology. He has
served as an Associate Editor of The Visual
Computer, Computers \& Graphics, and Computer
Graphics Forum. His primary research interests
include computer graphics and human-computer interaction.
\end{IEEEbiography}

%\vfill




\section*{Supplementary material}

# Supplemental Materials
# Sketch Beautification: Learning Part Beautification and Structure Refinement for Sketches of Man-made Objects

## 1 METHOD

We show our designed sketch implicit model and sketch assembly model in Figures 1 and 2, respectively.

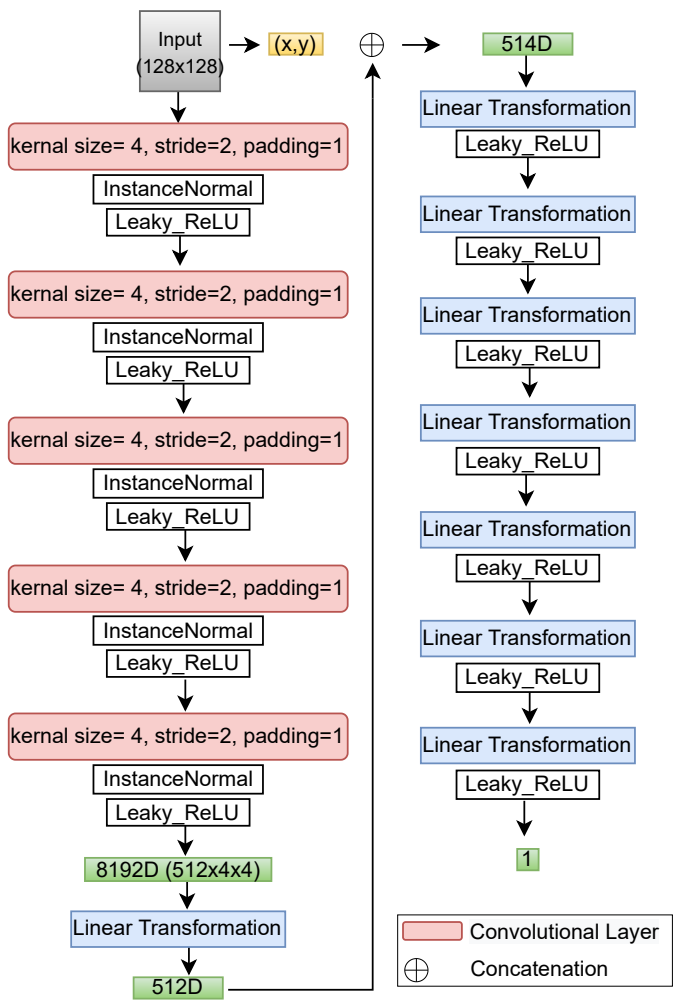

Fig. 1. Network structure of the sketch implicit model.

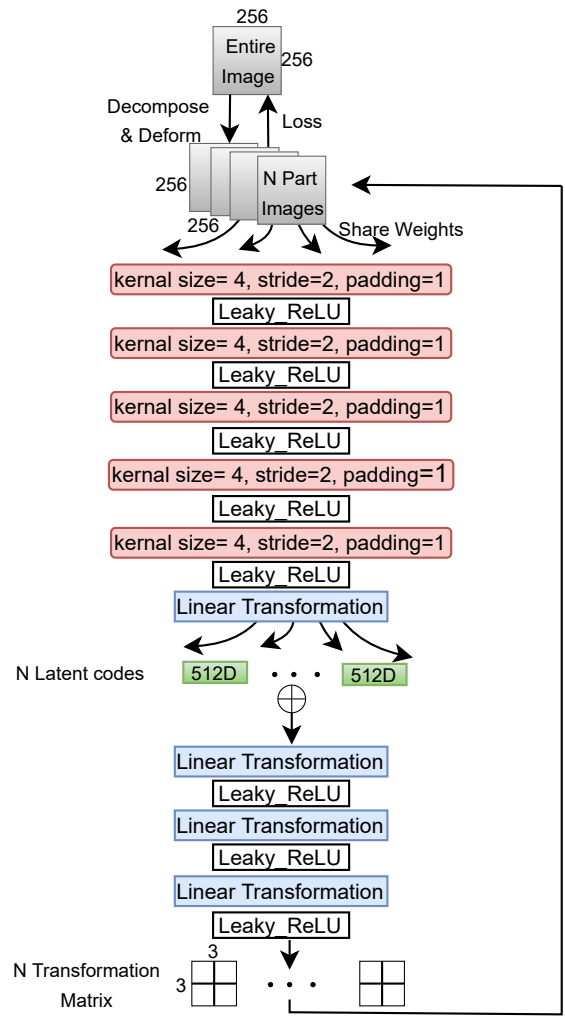

Fig. 2. Spatial transformation network (STN) for sketch assembly.

## 2 EXPERIMENTS

### 2.1 More Results

We show more results of our sketch beautification method in Figures 3 and 4 and 5.

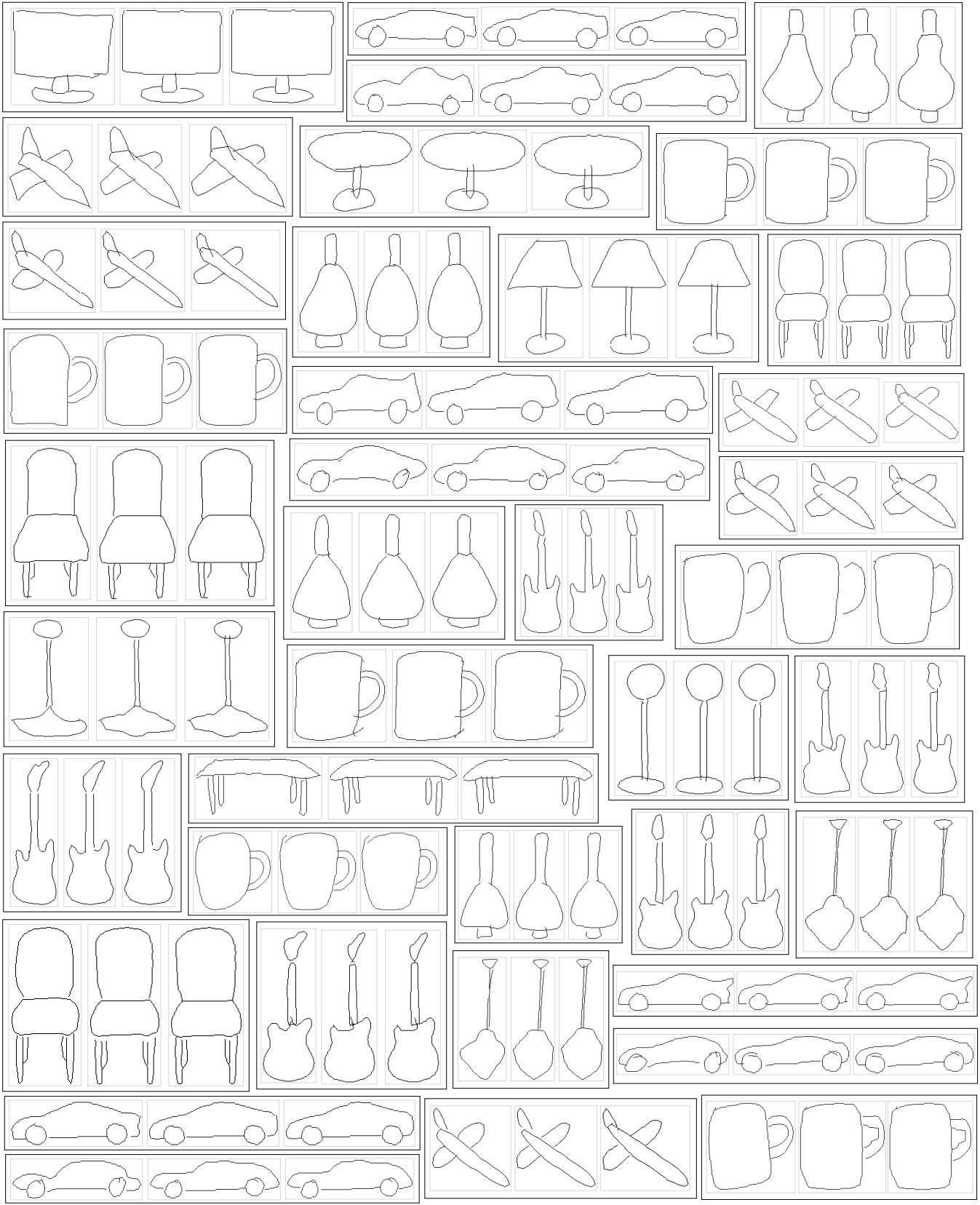

Fig. 3. Visual results of our method. Each triplet contains an input sketch (Left), the sketch after part beautification (Middle), and the final result after structure beautification (Right). Please zoom in for better visualization.

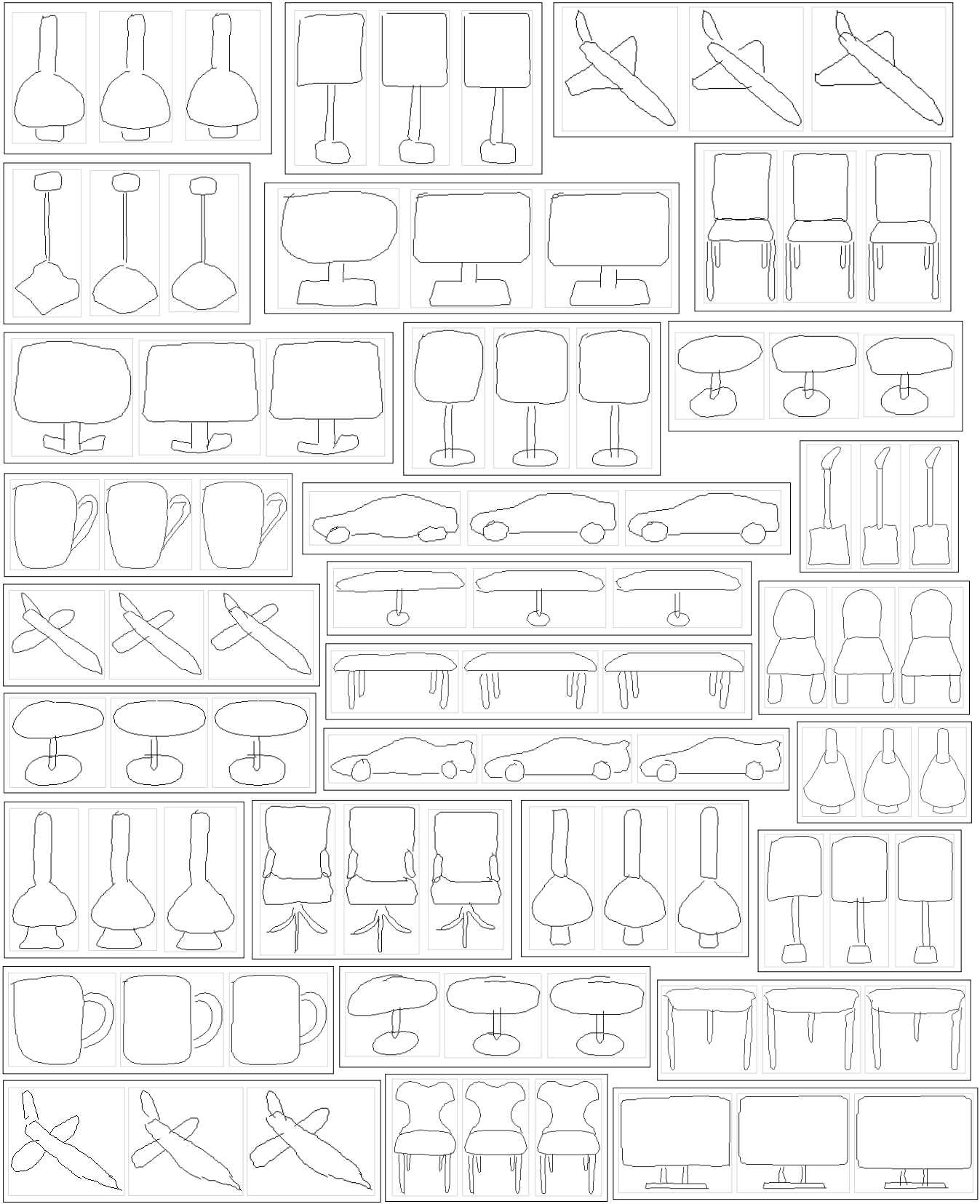

Fig. 4. Visual results of our method. Each triplet contains an input sketch (Left), the sketch after part beautification (Middle), and the final result after structure beautification (Right). Please zoom in for better visualization.

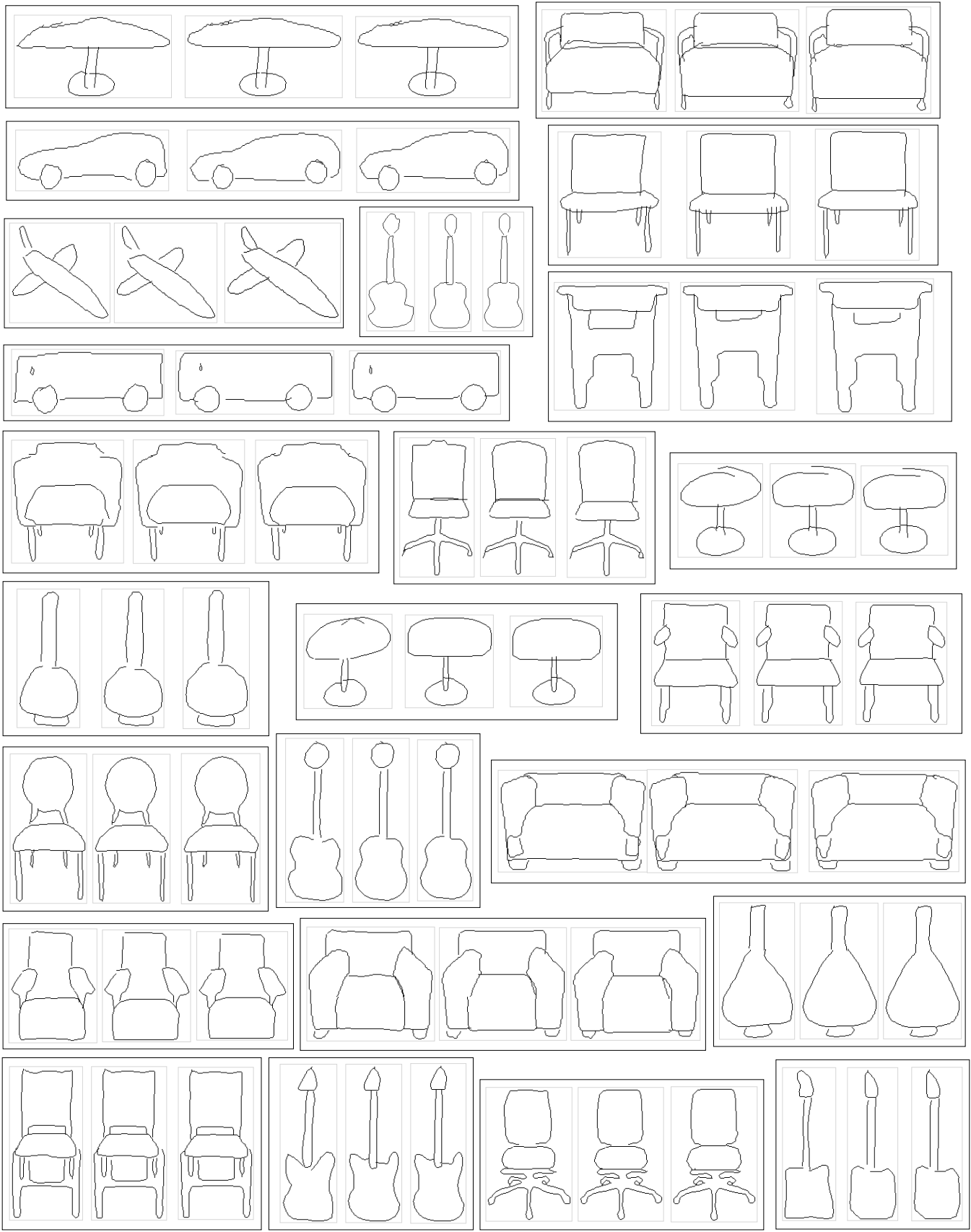

Fig. 5. Visual results of our method. Each triplet contains an input sketch (Left), the sketch after part beautification (Middle), and the final result after structure beautification (Right). Please zoom in for better visualization.

| Methods | mCD ↓ | mEMD(× $10^2$) ↓ |
|---|---|---|
| Part Retrieval (w/ HOG) | 8.22 | 5.20 |
| Part Retrieval (w/ Our Feature) | **5.91** | **4.82** |

TABLE 1
Quantitative evaluation of sketch beautification between part retrieval with the HOG descriptors and our trained sketch features.

## 2.2 Our Sketch Feature vs. HOG Descriptor

To demonstrate the effectiveness of our newly trained sketch feature codes in the sketch implicit model (Section 3.2.1), we compare our trained sketch feature codes with the HOG descriptors used in the retrieval-based methods, e.g., part retrieval. Table 1 illustrates the quantitative performance of the two methods. Figure 6 also shows the beautified results of part retrieval with the HOG descriptors and our trained sketch features. Compared to the HOG descriptors, our trained feature achieves better performance with closer distance to the users' freehand sketches under two metrics of the Chamfer distance and EMD distance.

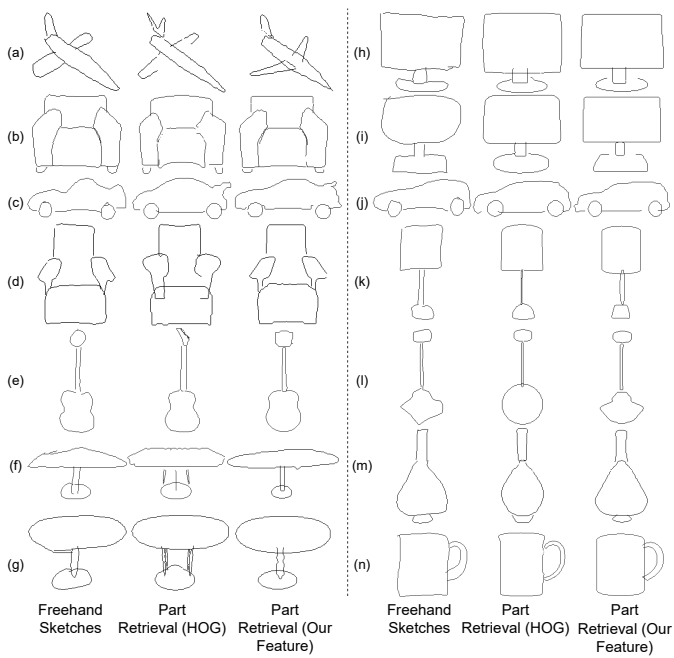

Fig. 6. Sketch beautification using part retrieval with the HOG descriptors and our newly trained sketch feature codes. Please zoom in for better visualization.

## 2.3 Our Sketch Beautification Method vs. Image-to-Image Translation Method

To demonstrate the effectiveness of our divide-and-combine strategy used in our sketch beautification method, we perform a comparative experiment where we treat a sketch differently as a holistic noisy (Figure 7) input with deformed parts (Section 3.3). Then, we try to solve the sketch beautification problem holistically as an image-to-image translation problem that takes a synthesized noisy sketch as input and produces a beautified sketch. Here, we compare the sketch assembly in our approach with a more direct image-to-image translation baseline (namely, pix2pix [1]) on the same dataset.

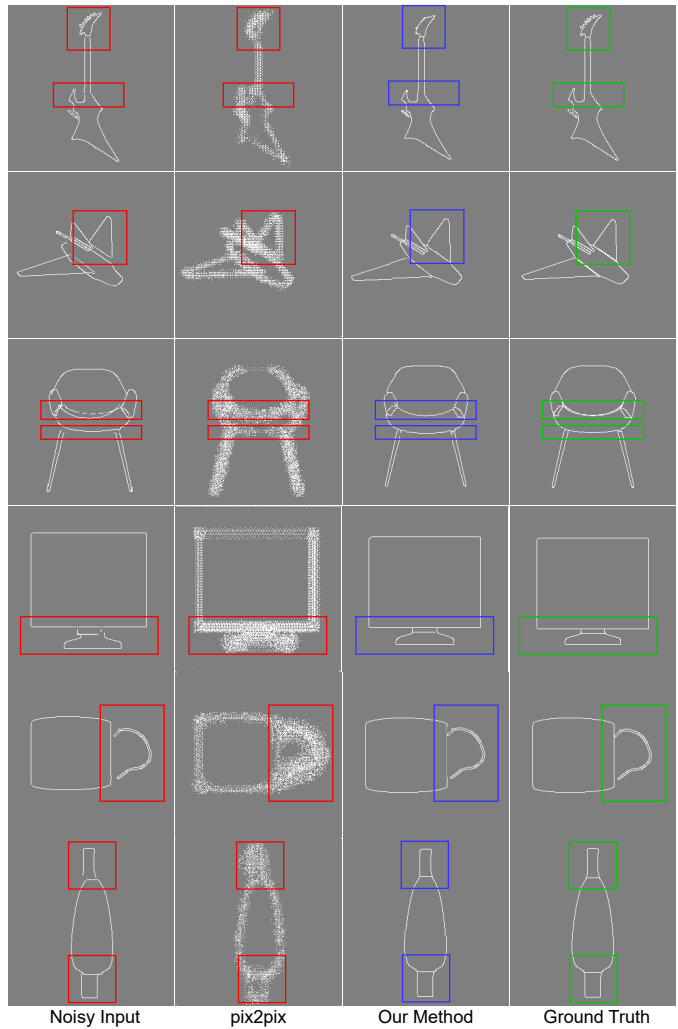

Fig. 7. Sketch beautification with pix2pix and our sketch assembly model. Please zoom in for better visualization.

Figure 7 shows that pix2pix tends to generate messy results with many noisy pixels and cannot beautify the structure defects (e.g., seams, misalignment, and penetrations between parts) precisely. Specifically, pix2pix usually heals the structure defects in the noisy input by generating random noise pixels surrounding those structure defects (red boxes in Figure 7). This is due to the fact that the combinations of deformations of different parts in a sketch are too complicated and cannot be easily solved by pix2pix simultaneously and holistically. Using our divide-and-combine strategy to compute the transformations for different individual parts in a sketch separately, our sketch assembly model achieves significantly better performance (blue boxes in Figure 7).

## REFERENCES

[1] P. Isola, J.-Y. Zhu, T. Zhou, and A. A. Efros, "Image-to-image translation with conditional adversarial networks," in *Proceedings of the IEEE conference on computer vision and pattern recognition*, 2017, pp. 1125–1134.



\end{document}

%% file: sections/abstract.tex
\begin{abstract}
 % background and introduction
We present a novel freehand sketch beautification method, which takes %we take
as input a freely drawn sketch of a man-made object and %our approach
automatically beautifies it both geometrically and structurally. 
% Challenging
Beautifying a sketch is challenging because of its highly abstract and heavily diverse drawing manner. 
% problem with existing method
{Existing methods are usually confined to 
their limited training samples} and thus cannot beautify %the
freely drawn sketches with {both geometric and structural variations}.
%motivation and our method
{To address this challenge, we}
adopt a divide-and-combine strategy{. Specifically, we} 
first parse {an} %the 
input sketch into {semantic components}, 
{beautify individual components by a learned} %a 
part beautification module {based on} part-level implicit manifolds{, and
then reassemble the beautified components through a structure beautification module.}
With this strategy, our method can go beyond the training samples and handle novel freehand sketches. 
%evaluation
We demonstrate the effectiveness of our system with extensive experiments
and a {perceptual} study.

%-----------------------------------------------
\end{abstract}  
%-------------------------------------------------------------------------

%% file: sections/introduction.tex
\section{Introduction}
\label{section:introduction}
%%background
\IEEEPARstart{S}{ketching} is a universal and intuitive tool for humans to render and interpret the visual world.
Even {if} sketches are usually drawn in an imprecise and abstract format, human viewers can still {implicitly} %automatically 
beautify them and {easily} envision their underlying objects. % easily.
But for machines, existing computer algorithms \cite{delanoy20183d,lun20173d} are struggling to make use of these freely drawn sketches directly, in particular, the sketches created for {depicting} man-made objects with diverse geometry and non-trivial topology.
Although the beautification problem has been studied for decades{,} ranging from the primitives of %the 
sketched geometric objects \cite{wang2005exploring} to strokes in handwritings \cite{zitnick2013handwriting}{, systematic} %.     Systematic 
analysis of beautifying sketches for man-made objects has rarely been studied. % yet.
The key challenge {here} 
is how to instantiate %the 
poorly drawn conceptual geometries and {refine imprecise structures} 
simultaneously.
Addressing the beautification problem of %the
man-made object sketches can inspire and facilitate {various} %many 
downstream sketch-based applications such as sketch-based modeling \cite{smirnov2020learning, delanoy20183d,su2017sketch2normal}, sketch-based retrieval \cite{chen2018deep, dai2017deep}, and other sketch understanding task{s} \cite{yu2020sketchdesc}.

%% research progress and their problem
As sketches are widely utilized in manufacture designing \cite{li2020sketch2cad, gryaditskaya2019opensketch, li2022free2cad}, animation drafting \cite{su2018live, simo2018mastering} and HCI (Human-Computer Interaction) \cite{ye20213d}, 
many algorithms {have been} %are 
presented to process %the 
freely drawn sketches targeting vectorization \cite{orbay2011beautification}, rough sketch cleanup \cite{simo2018mastering}, and simplification \cite{van2021strokestrip, liu2018strokeaggregator}. 
These methods are common in two places: {first,} %one is that
they usually produce more local modifications on input sketches, and seldom consider or touch the global structures; {also,} %two is that
the input of these methods is nearly ready-to-use sketches with perfect strokes that are straight, mutual{ly} parallel {or} %and 
curved with perpendicular angles.
Hence, the aforementioned approaches are inapplicable to our task of beautifying the artifacts in the man-made object sketches with diverse local distortion and global inconsistency.
As for
{the drawing assistance to}
% on 
the overall structures, prior approaches \cite{bae2008ilovesketch, dixon2010icandraw, lee2011shadowdraw} are mainly designed in a heuristic and {interactive} way that {provides a holistic scaffold or shadow guidance, and 
updates their guidance based on drawn strokes} 
during the drawing process.
{Alternatively, Fi{\v{s}}er et al. \cite{fivser2016advanced} proposed a rule-based stroke beautification approach{, which, however, is} % but 
still conditioned on the former strokes and confined to the stroke sequences.}
{Therefore, these methods} cannot be applied as post-processing to the existing hand-drawn sketches {without the drawing order of individual strokes}.

\begin{figure*}[tb]
 \centering
 \includegraphics[width=\linewidth]{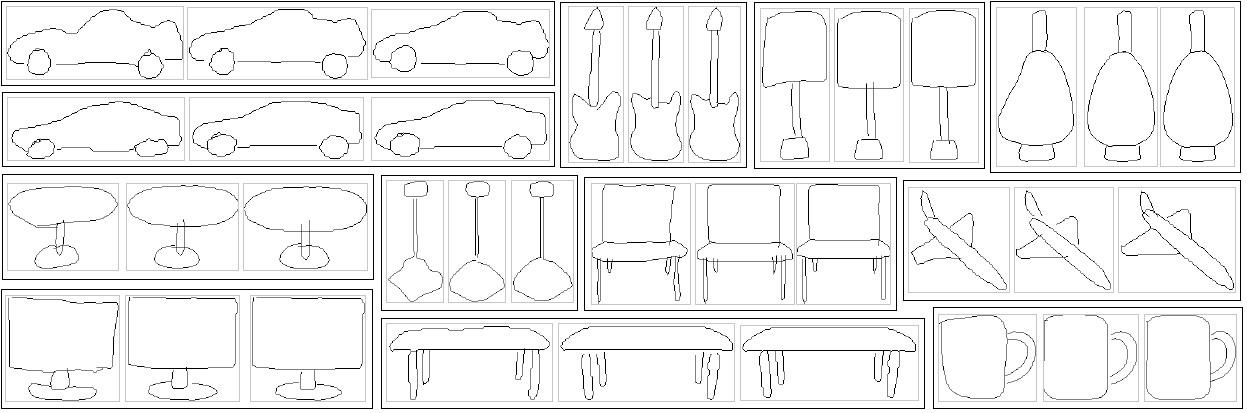}
  \caption{We present a novel technique for beautifying freehand %Beautification of
  sketches of man-made objects. Each triplet contains {an input} %user's freehand
  sketch (left), the sketch after part beautification (middle), % part beautified sketch,
  and the final result {after structure refinement} (right).}
    \label{fig:teaser}
\end{figure*}

    %%observation 1
    %%observation 1
   
    {We observe that the traditional data-driven methods \cite{simo2018mastering, simo2016learning} usually lack robustness and fail to produce satisfactory beautification results for freely drawn sketches of man-made objects, especially when the inputs are created with large variations.}
    This is because their trained inference models are heavily confined to 
    % the distribution of
    {the limited training samples}. 
    We are largely inspired by CompoNet\cite{schor2019componet}, which {transcends}
    % jumps out of % where they jumped 
    the bound of the empirical distribution of the observed data and {enriches} %enriched 
    the distribution diversity of generated samples by learning both novel {part} % parts 
    synthesis and plausible part compositions.
    {Different from their generation task that tries to synthesize infinite outputs from finite inputs, our beautification task can be regarded as {the reverse} process that aims to beautify the infinite freehand inputs to the finite outputs.}

    %% our idea and method
    In this work, we treat {a sketched object as} %the entire sketch as 
    a combination of %various 
    part sketches and further transform the 
    {sketch} %sketches
    beautification problem into two sub-problems:
    part sketch beautification and 
    {global structure beautification}.
    {Given a freehand input sketch, we expect to beautify the local geometry defects in the part beautification module and address the {overall} structure issues in the
    % part assembly module.
    {structure beautification module.}
    To better instantiate the user's conceptual sketches, we {make} %made
    several strict constraints for this task.
    First, during the process of part beautification,  one should follow {the} user's original drawing intention as much as possible.
    {In our algorithm, we regard this constraint as: (i) The endpoints of the strokes in the beautified result should be as close as possible to
    % remain the same
    {those} in the user's original sketch;
    (ii) The curvature of the strokes in the beautified result should be as close as possible to its beautification reference (see Figure \ref{fig:geometry_beautification}).}
    Second, during the process of
    {structure beautification}% part assembly
    , one should maintain the user's original structure as much as possible.
    Therefore, directly replacing the user's input part sketch with a 
    % perfect 
    {better
    part sketch retrieved from the database} 
    is not considered in our method.
    In addition, we do not aim for large transformations and structural {alterations} {(e.g., scaling beyond
    % by more than 
    0.8 $\sim$ 1.2 times or translating beyond
    % by more than
    $-3$ to $3$ 
    pixels on the individual parts of a sketch)}, and we perform
    %a mild
    smaller refinements on {the input sketch in terms of both geometry and structure} 
    (see Figure \ref{fig:teaser}).  }
    %method

\begin{figure}[tb]
    \centering
        % the following command controls the width of the embedded PS file
        % (relative to the width of the current column)
        \includegraphics[width=\linewidth]{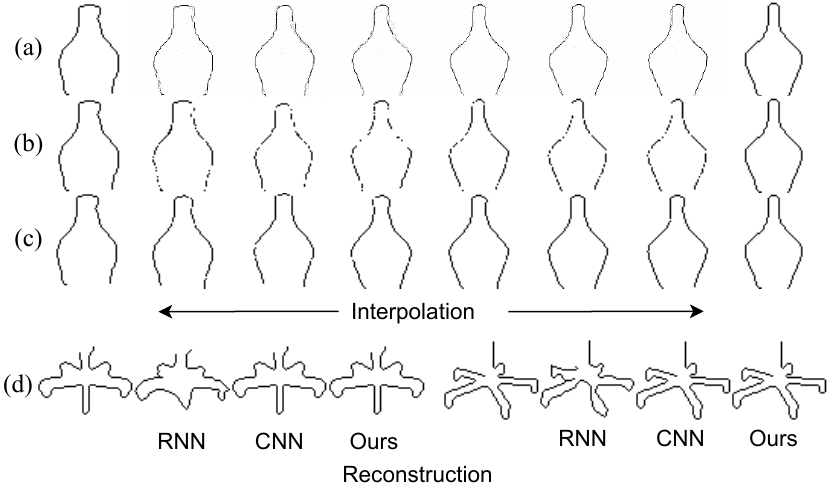}
        % replacing the above command with the one below will explicitly set
        % the bounding box of the PS figure to the rectangle (xl,yl),(xh,yh).
        % It will also prevent LaTeX from reading the PS file to determine
        % the bounding box (i.e., it will speed up the compilation process)
        % \includegraphics[width=.95\linewidth, bb=39 696 126 756]{sampleFig}
        %
        % \parbox[t]{.9\columnwidth}{\relax    
        % Please zoom in for a better visualization.}
        %
        \caption{\label{fig:intepolation_compare}
                 Sketch interpolation {based on} % of 
                 CNN ({a}), {CNN after binarization with the threshold of $0.5$ (b),} and our implicit representation ({c}), and reconstruction with different representations ({d}). Please zoom in to {examine the details{, e.g., {blurry} %blurs 
                 and diffusion artifacts surrounding the interpolated strokes
                 based on CNN (a), as well as {broken} %the breaking
                 strokes and missing points with CNN after binarization (b).}} }
      \end{figure}

    In the part beautification stage, we argue that existing {deep} 
    sketch representations ({i.e.,} 
    CNN features from rasterized  {pixel maps} 
    \cite{yu2015sketch} or RNN features from stroke sequences \cite{ha2018neural}) are insufficient to perform beautification on users' conceptual freehand sketches, {as shown in} %see 
    Figure \ref{fig:intepolation_compare}.
    %It is known that
    CNN representations are widely utilized in natural image synthesis and can easily generate plausible and novel samples by interpolating existing images \cite{wu2019transgaga} or the disentangled semantic attributes \cite{karras2020analyzing}. However, different from {natural} %the natural 
    images, sketches are usually too sparse with few valid elements or points that cannot be smoothly interpolated with CNN representations {(see the comparison of sketch interpolation in Figure \ref{fig:intepolation_compare})}.
    As for RNN representations, existing methods cannot guarantee a precise reconstruction of %the 
    input sketches {(see the comparison of sketch reconstruction in Figure \ref{fig:intepolation_compare})}, not to mention the {more challenging interpolation task}. %later stage.
    In this paper, we propose a novel implicit sketch representation{, which} % that 
    is not only able to represent sketches effectively but also can construct a continuous and smooth manifold to synthesize and instantiate users' conceptual drawings by interpolating the {existing sketches}.

    In the 
    % part assembly
    {structure beautification}
    stage, to simulate the structural {errors} 
    {that exist in the user's input sketches}, 
    we apply random affine {transformations} to the ground-truth sketches in our training dataset and enforce our
    % part assembly module
    {designed sketch assembly model}
    to learn these transformations between the {transformed} %affined 
    part sketches and their ground truth counterparts. 
    Then, simply feeding the input sketches to our 
    % part assembly module
    {sketch assembly model}
    usually causes learning 
    % vibration
    {oscillation}
    and failures in model convergence due to the sparsity of the input sketches.
    To address this issue, we use part-level bounding boxes to enhance the spatial feature of the separate part sketches{, where the part bounding boxes (Figure \ref{fig:fig_part_assembly_pipeline} (e)) are used to specify the spatial location of individual part sketches and the spatial relationship between different part sketches in a user's drawing}.
    We propose an 
    {IOU} {(intersection over union)} metric {{which uses} 
    the overlapping between the bounding boxes of transformed part sketches and their corresponding part ground truth}
    {to} %is 
    further 
    evaluate the performance of the %part assembly module.
    {sketch assembly model}.
    
    Training our beautification framework requires a {considerably} %considerable
    large amount of part-annotated 
    sketch data. Considering the limited training sketches, we collect a novel {training dataset of} part-labeled sketches % training dataset 
    by rendering the edge map{s} with the semantic labels under the best view \cite{dutagaci2010benchmark} from the {existing} %current
    3D shape repositories, 
    i.e., PartNet \cite{mo2019partnet}, SDM-Net \cite{gao2019sdm}, and COSEG dataset \cite{sidi2011unsupervised}. 
    %% evaluation discussion
    {We evaluate the beautification results in terms of faithfulness to the input sketch and the beautification quality}.
    {Intuitively, faithfulness can be evaluated quantitatively and via a user study, while the beautification quality can only be evaluated via a user study due to its subjectiveness.
    In this work, we perform a quantitative evaluation on faithfulness and a 
    {perceptual} % perceptive
    study on beautification quality to interpret the performance of sketch beautification methods more comprehensively.   
    }

    %% contributions
    Our contributions in this work are:
    \begin{itemize}
    \item We introduce a novel beautification framework for sketches of man-made objects, which consists of % containing
    a part beautification module and a %part assembly
    {structure beautification}
    module.
    \item We propose a novel sketch implicit representation that can represent sketches {effectively}. 
    {Converting the input freehand sketches to implicit representations is the primary and essential step in our part beautification stage.} To the best of our knowledge, we are the first to represent sketches with implicit functions. 
    \item We design a novel sketch assembly
    %module
    {model} to learn the spatial transformations for the sparse representations {to achieve structure beautification}.
    \item We collect a large-scale part-labeled sketch dataset consisting of 17,172 sketches spanning 9 object categories{. We will release the data and code} %, and will release the dataset 
    to the research community.
    \end{itemize}

%% file: sections/related_work.tex
\section{Related Work}
In this section, we review the previous approaches that are closely related to our work in these categories: sketch beautification, sketch representation{s}, and implicit representation{s}.

\subsection{Sketch Beautification}
\label{subsection: Sketch Beautification}

In the past decades, numerous algorithms were proposed to beautify %users'
freehand drawings. {The}
%At the outset, the 
earliest algorithmic beautification method with aesthetic constraints dates back to SketchPad \cite{sutherland1964sketchpad}. 
Later, Pavlidis and Van Wyk \cite{pavlidis1985automatic} %formulated the problem statement and 
proposed an algorithm for beautifying figures as a post-process.
{{Igarashi} %Takeo
et al. \cite{igarashi2007interactive} presented a system for rapid geometric design that beautified {strokes} %the user's stroke 
in an interactive way.}
Recently, great strides have been made in the sketch simplification \cite{van2021strokestrip, liu2018strokeaggregator}, rough sketch {cleanup}
\cite{simo2018mastering} and sketch vectorization \cite{orbay2011beautification}.
Since these methods do not attempt to change {or beautify} the global structure of %the 
input sketches, we do not conduct a detailed review here.
We refer the interested readers to \cite{yan2020benchmark} for an insightful survey.

The more related works are iCanDraw \cite{dixon2010icandraw}, ShadowDraw \cite{lee2011shadowdraw}, and ShipShape \cite{fivser2016advanced}.
{%For 
iCanDraw and ShadowDraw %, this two approaches 
can output a sketch with {a} reasonable layout and more realistic details by providing reference scaffolds or shadow layouts for %the
users.}
However, as discussed before, limited by their heuristic and iterative settings during drawing,  
{they}
cannot beautify the existing sketches. 
{For ShipShape, this method}
% ShipShape 
only performed a curve{-}level beautification in {an} %a
interactive way by enforcing some geometric constraints to the afterwards curves based on the previous strokes, such as symmetry, reflection, and arc fitting.
These constraints are difficult to apply to the existing sketches in practice without knowing the drawing order of the component strokes. 
{Besides, during its beautification process, %the
extra efforts from users on the structure adjustment are still needed.} {Instead}, %While
we %instead
study the part-level beautification %here 
since the part labels are more accessible with existing segment approaches \cite{huang2014data,li2018fast, yang2021sketchgnn} or {user-specified} %user's 
annotations. {We further {reduce} %free
user labors with our 
% part assembly module
{sketch assembly model}
by performing structure beautification automatically.}

A similar beautification problem was also explored in EZ-Sketching, where Su {et} %et.
al. \cite{su2014ez} proposed a three-level optimization framework to beautify {a} %the
sketch traced over {an} %the 
image.
{Their beautification process heavily relies on the image being traced, which is missing in our task.}
In DeepFaceDrawing, Chen et al. \cite{chen2020deepfacedrawing} presented a robust portrait image synthesis method{, which can handle rough or incomplete} %with rough input
sketches as input.
By {decomposing a face sketch into face components and} projecting the face component sketch{es} to the {component-level} %part-level 
sketch manifold{s}, the {input} %user's
freehand sketches can be {implicitly refined for generating photo-realistic face images}. 
As the layouts or structures of human portraits are almost fixed, their approach is not suitable for our %man-made object sketch
beautification task.
To handle the diverse structure of %the
sketched man-made objects, we propose a novel sketch 
% part 
assembly module, as inspired by \cite{schor2019componet}.
 \begin{figure*}[htb]
      \centering
      % the following command controls the width of the embedded PS file
      % (relative to the width of the current column)
      \includegraphics[width=\linewidth]{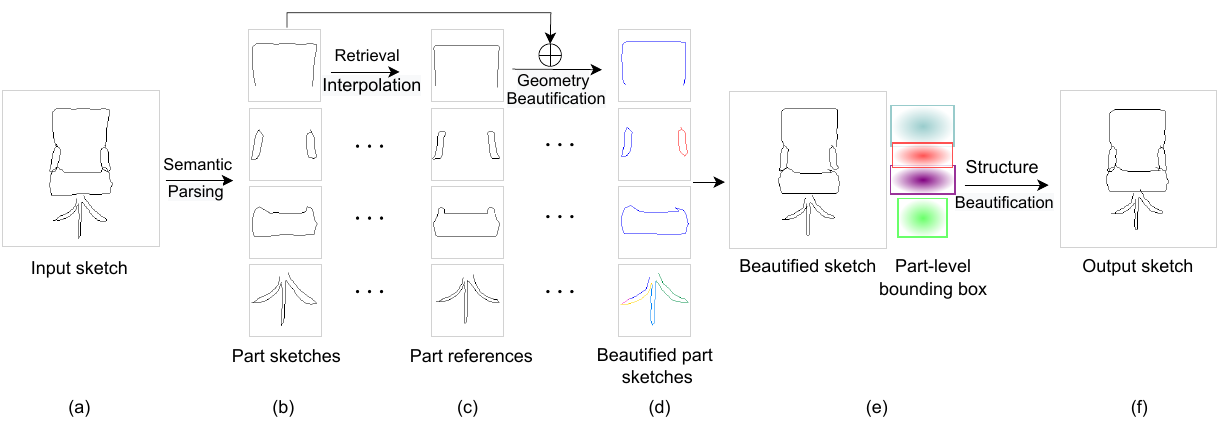}
      % replacing the above command with the one below will explicitly set
      % the bounding box of the PS figure to the rectangle (xl,yl),(xh,yh).
      % It will also prevent LaTeX from reading the PS file to determine
      % the bounding box (i.e., it will speed up the compilation process)
      % \includegraphics[width=.95\linewidth, bb=39 696 126 756]{sampleFig}
      %
     % \parbox[t]{.9\columnwidth}{\relax}
      %
      \caption{\label{fig:fig_pipeline_sketch_beautification}
             {System} pipeline of sketch beautification.
             {For an input sketch (a), we first parse it to individual part sketches (b), {and} %we can 
             then synthesize the corresponding part references (c) by retrieval and interpolation. 
             After performing geometry beautification on the part sketches (b) towards part references (c), we %can
             obtain the new part sketches (d) with well-beautified geometry (see geometry differences between (b) and (d)). 
             During the stage of structure beautification, we adjust the imperfect structure (notice the misalignment of chair arm{s}) of the intermediate output (e) with the help of part-level bounding boxes (e) and generate the final {beautified sketch} %sketch beautification 
             (f). % for the input sketch(a). 
             Different colors in beautified part sketches (d) indicate the different strokes.
             {The} %And 
             colorful bounding boxes (e) denote the scales and spatial locations of different part sketches in the image space (256 $\times$ 256).}}
    \end{figure*}

\subsection{Sketch Representations}
A sketch is generally represented as either a rasterized binary-pixel map \cite{schneider2014sketch} or vector sequences \cite{ha2018neural}, {\cite{ribeiro2020sketchformer}}, or both \cite{bhunia2021vectorization}.
Since the sketch datasets TU-berlin \cite{eitz2012humans} and Quickdraw \cite{ha2018neural} were introduced, the research community has seen many works in sketch understanding. {For example},
Yu et al. \cite{yu2015sketch} proposed a multi-scale and multi-channel CNN framework with a larger kernel size for %the 
sparse raster sketches.
Ha and Erc \cite{ha2018neural} introduced a sequence-to-sequence variational auto-encoder with the bidirectional RNN backbone for %the 
vectorized {sketches}. %input.
Later, the CNN and RNN representations \cite{xu2018sketchmate, li2020sketch} were combined to better represent sketches in their retrieval and recognition tasks, respectively. {Most of the subsequent}
%Following 
works {\cite{bhunia2021vectorization,bhunia2022sketching}} 
basically inherited and employed the off-the-shelf backbones for the specified tasks.
However, {since the} %as
CNN-based representations are designed to learn the distribution of all the pixels in the 2D image space rather than the solo valid points on %the 
strokes, they tend to generate {broken} %breaking
segments with shadow effects (Figure \ref{fig:intepolation_compare} {(a) and (b))}
during the interpolation process.
%And
The RNN-based representations failed to reconstruct input sketches with vector sequences precisely (Figure \ref{fig:intepolation_compare} {(d)}). 
{Alternatively, Sketchformer\cite{ribeiro2020sketchformer}  proposed a transformer-based representation for encoding {a freehand} %the free-hand 
sketch input in a vector form. 
Despite its success in generating more plausible sketch interpolation compared to RNN-based representations, this method still fails to faithfully reconstruct %the 
complicated part sketches (e.g., chair legs) and tends to produce shifting issues during interpolation due to their stored relative position for sketch points in their vectorized format of input sketches.}
Therefore, we turn to implicit functions to better represent part sketches in this work.

\subsection{Implicit Representations}
Recently, implicit functions have attracted extensive attention in the research community \cite{deng2021deformed, chen2020bsp, saito2019pifu, park2019deepsdf, mescheder2019occupancy, chen2019learning}{, since they}
can represent 3D shapes in a continuous and smooth implicit field.
Existing implicit representations typically used a spatial function to represent {a shape} %the shapes 
by mapping the inside and outside points distinguishably.
In {DeepSDF}, % Deepsdf,
Park et al. \cite{park2019deepsdf} utilized {a} %the 
signed distance function {(SDF)} to represent a watertight shape in 3D space where inside and outside points are {respectively} mapped to negative and positive values, and the underlying surface is implicitly represented by {a} % the 
zero-crossing surface.
Similarly, Mescheder et al. \cite{mescheder2019occupancy} represented  {a} %the 
target shape with a continuous occupancy function, indicating the probability of points being occupied by a shape.
However, different from %the 
{watertight }3D shapes, sketches are {often} created with no concept of \textit{inside} or \textit{outside} and are usually too sparse with limited valid points or pixels in the 2D image space. Hence, the aforementioned implicit representations 
cannot {be directly applied} %directly apply
to the part sketches. 
{To address this issue, we design an intuitive 2D implicit function (we call it a hit function) for part sketches, where we map all sampled points of a canonical 2D space to two exclusive statuses: whether its projection to the image space of the part sketch %does
\textit{hits} the stroke or {not} % \textit{not hit} the stroke 
(see Figure \ref{fig:fig_sketch_implicit_representation}).}

%% file: sections/methodology.tex
\section{Methodology}
In this paper, we mainly focus on %the 
freehand sketches created for man-made objects {that can be segmented into parts}. 
We treat this kind of sketch{es} as a combination of 
%the 
individual part sketches{, since 3D man-made objects often consist of semantically meaningful parts}
{\cite{li2017grass,wang2018global,wu2020pq}}.
{Our} %The 
framework for sketch beautification is illustrated in Figure \ref{fig:fig_pipeline_sketch_beautification}. 
We first parse {an input} 
%the 
freehand sketch {depicting a man-made object like a chair} to part-level sketches, {and} then 
%we 
beautify the individual part sketches in the part beautification module. In this module, we construct {an} 
%a
implicit manifold for each part. {For each part, we can} %where the part sketch can 
search and {incorporate} the geometry features from its key neighbors in this manifold {for geometry beautification}.
{``{Incorporating}'' the geometry features happens in the
linear interpolation (Section \ref{subsubsection: Retrieval_and_Interpolation})
and part geometry beautification (Section \ref{subsubsection:geometry_beautification}).} 
Finally, the beautified part sketch{es are} 
%will be 
recomposed as the final beautified sketch by 
% our part assembly module
{our sketch assembly model}, 
{which performs the structure beautification task}.

\begin{figure}[tb]
      \centering
      % the following command controls the width of the embedded PS file
      % (relative to the width of the current column)
      \includegraphics[width=\linewidth]{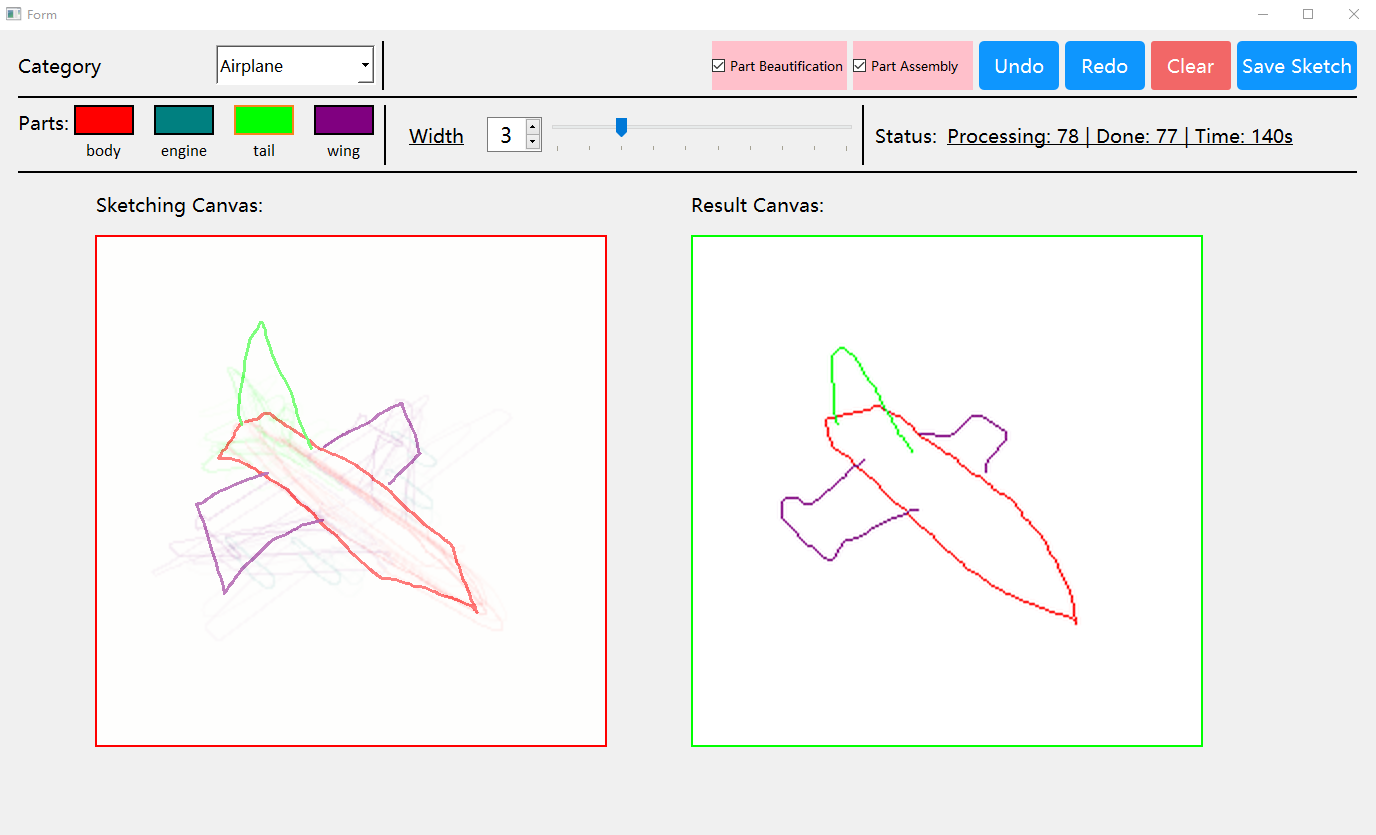}
      % replacing the above command with the one below will explicitly set
      % the bounding box of the PS figure to the rectangle (xl,yl),(xh,yh).
      % It will also prevent LaTeX from reading the PS file to determine
      % the bounding box (i.e., it will speed up the compilation process)
      % \includegraphics[width=.95\linewidth, bb=39 696 126 756]{sampleFig}
      %
     % \parbox[t]{.9\columnwidth}{\relax} deng adding some illustration.
      %
      \caption{\label{fig:fig_interface}
               {Our sketching} %Sketch 
               interface. }
    \end{figure}
\subsection{{Sketching} Interface}
{Our algorithm requires semantic segmentation of an input sketch. Sketch segmentation can be done automatically, as demonstrated in previous works {\cite{huang2014data,li2018fast, yang2021sketchgnn}}. However, for simplicity, we design a simple interface (Figure \ref{fig:fig_interface}) for users to interactively provide part-level semantic information while sketching.}
The key {enablers} %components 
of this system are the part beautification {component} and the 
% part assembly
{structure beautification}
{component}. Once activated on the interface {(through ``Part {Beautification}" and 
% ``Part Assembly''
{``Structure Beautification''}
buttons)}, 
{the} two functions will run in the background and beautify the user's input {automatically after the user finishes one complete sketch.}
We provide more details of the mechanism of these two functions in Sections \ref{subsection: part_beautification} and \ref{subsection: part_assembly}.

{Before sketching, a user first needs to select a target object category from one of our prepared nine classes (e.g., airplane, chair, table).
Our interface then shows the corresponding set of part labels. It also provides part-level shadows to provide initial geometry and structure guidance, similar to ShadowDraw \cite{lee2011shadowdraw}. Unlike ShadowDraw, we}
do not update the underlying shadows dynamically during the drawing process {since} %because 
we hope to give more freedom to users and not influence them too much.
In this interface, we provide several basic drawing tools for users to amend their drawings, such as clear, undo, redo, etc.
By clicking on a part label (e.g., airplane engine), the strokes drawn afterwards will be labeled with the selected label automatically.
In this way, we obtain a freehand sketch with well-segmented parts.
Note that our system saves {each drawn} %the resulting 
sketch in both {raster and vector graphics formats}. 
For {a} % the 
better drawing experience, our interface {supports the drawing of strokes with varying 
% thickness
{thicknesses}. However, such strokes are pre-processed {by extracting the skeletons  
{\cite{zhang1984fast}} from the user's sketch input to} have one-pixel width for further processing.}
{Note that we utilize the “skeletonize” operation for real-time processing of %the user’s 
freehand sketches when implementing our sketching interface,  rather than directly rasterizing the recorded vectorized stroke sequences {since} %due to the fact that 
{the latter's performance highly depends on the number of input strokes}.} 

\subsection{Part Beautification}
\label{subsection: part_beautification}
Given a rasterized freehand sketch with part annotations, we treat its component 
sketches separately and beautify {them individually} 
in the corresponding part-level implicit manifold{s}. Different from CNN representations created for the dense and high-frequency RGB pixels, {our novel} %we propose a novel 
implicit representation {works} for the low-frequency and sparse sketched points in {the} 2D image space.
{Existing} %As existing 
implicit representations are not suitable for our sketch points in the 2D space, {since a 2D sketch is a kind of more discrete representation having no inside and outside spaces and occupying no continuous regions compared with the closed 3D shapes}. 
We {thus} design our own sketch implicit representation {where we first sample points from a 2D canonical space{. We then} %, then 
project these points to the same-size image space of sketches and record whether these points could hit any stroke of a sketch{. Finally,} %; finally,
a discrete sketch can be represented by the sampled points (hitting sketches) of this continuous 2D canonical field implicitly (see Figure \ref{fig:fig_sketch_implicit_representation}).}

\subsubsection{Sketch Implicit Representation}
\begin{figure}[tb]
   \centering
   % the following command controls the width of the embedded PS file
   % (relative to the width of the current column)
   \includegraphics[width=.98\linewidth]{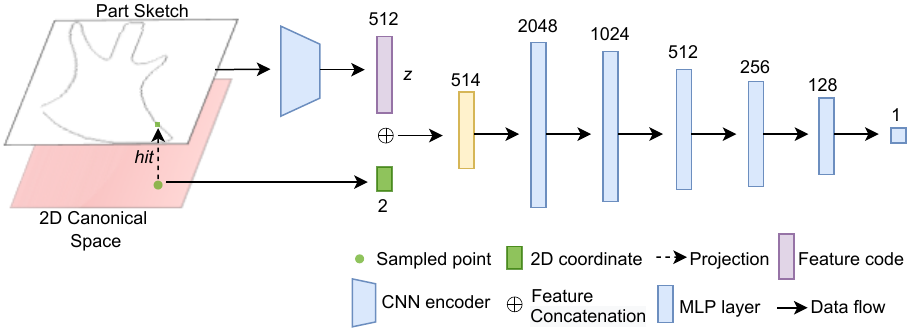}
   % replacing the above command with the one below will explicitly set
   % the bounding box of the PS figure to the rectangle (xl,yl),(xh,yh).
   % It will also prevent LaTeX from reading the PS file to determine
   % the bounding box (i.e., it will speed up the compilation process)
   % \includegraphics[width=.95\linewidth, bb=39 696 126 756]{sampleFig}
   %
  % \parbox[t]{.9\columnwidth}{\relax} deng adding some illustration.
   %
   \caption{\label{fig:fig_sketch_implicit_representation}
            {Illustration of our} pipeline for learning {a} sketch implicit representation.}
\end{figure}

An implicit field is typically defined by a continuous function over 2D/3D space.
In this work, we take {a sketch as an implicit function} 
({hit} function{)} 
defined over the continuous coordinates $P$ in a {2D canonical space}, as given in Equation \ref{eq:sk_im_define}.
\begin{equation}
   \label{eq:sk_im_define}
   H(p) =\left\{
   \begin{aligned}
   1 & , & \textrm{ \textit{p}{'s projection hits a stroke}},\\
   0 & , & \textrm{otherwise}.
   \end{aligned}
   \right.
\end{equation}
Here, we {define} 
the valid points (those projections that hit a stroke in a sketch) as 
{1}, and the {invalid} points (those projections that hit the empty background) as {0}. 
The underlying sketch is represented by the points with positive values, that is {$H(\cdot) = 1$}, 
% instead of
{similar to}
the zero-isosurface points in 3D implicit representations. 
We then employ a neural network $f_{\theta}$ to approximate the implicit function {$H(p)$} as  $f_{\theta}(p)$. 
{Inspired by \cite{chen2019learning}, where Chen and Zhang proposed an implicit decoder (IM-NET) to map a shape feature vector and a point coordinate to a value indicating the inside/outside status of {a} %the 
point relative to the shape, we adapt IM-NET to our sketch implicit function to indicate the status (hits a stroke or not) of a sampled 2D point (from the 2D canonical space) concatenated with a feature code $z$ of an input sketch (see Figure \ref{fig:fig_sketch_implicit_representation}). As for the sketch encoder, we design a} CNN encoder {(see {the} supplemental materials)}
% is further added
to extract the feature code $z$ from the input sketch. Conditioned on the feature code, the implicit function can be further formulated as $f_{\theta}(z,p)$. 

Here, we adopt 
% a
Multi-layer Perceptrons (MLPs) 
with rectified linear unit (ReLU) as the implicit function $f_{\theta}$. 
To represent part sketches in a uniform 2D space, we center-crop and resize all the part sketches to a $128 \times 128$ scale.
We {illustrate} % demonstrate 
the model for learning {our} sketch implicit representation in Figure \ref{fig:fig_sketch_implicit_representation}. The part sample is first encoded to %the 
{a 512D} %512-d 
feature code and then a sampled {2D} %2-d 
coordinate is concatenated 
with the {feature} code as the input to {the} implicit model. 
After the model converges, {an} %a 
input (part) sketch can be represented as the implicit representation $z$ and interpreted back to {the} 2D image space by sampling points to $f_{\theta}$.

\textit{Loss function.} {We use a mean squared error between ground truth labels and predicted labels for each point as the loss function}
%The loss function 
for training sketch implicit function $f_{\theta}$. % is a mean squared error between ground truth labels and predicted labels for each point. 
{Specifically,} we formulate it as follow{s}:
\begin{equation}
   \label{eq:sk_im_loss}
   L(\theta)={\sum}_{p \in P} {\lVert f_{\theta}(p) - H(p) \rVert}^2,
\end{equation}
where $P$ refers to the point set sampled from the $128 \times 128$ 2D space and {$H(p)$} is the ground truth value in our dataset given the query point $p$.

\subsubsection{Retrieval and % i
{I}nterpolation}
\label{subsubsection: Retrieval_and_Interpolation}
\begin{figure}[t]
   \centering
   % the following command controls the width of the embedded PS file
   % (relative to the width of the current column)
   \includegraphics[width=.96\linewidth]{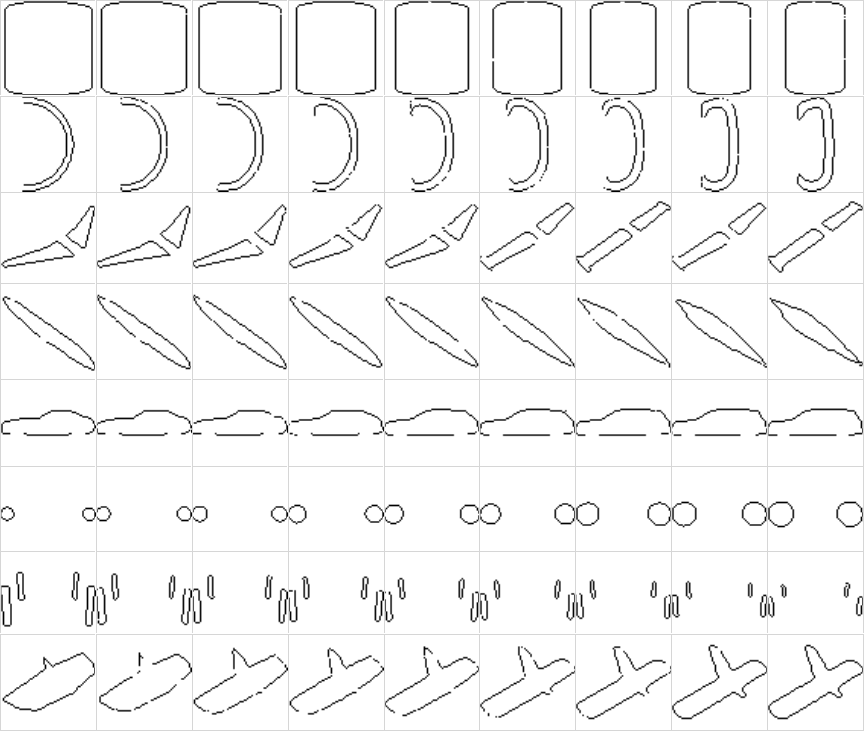}
   % replacing the above command with the one below will explicitly set
   % the bounding box of the PS figure to the rectangle (xl,yl),(xh,yh).
   % It will also prevent LaTeX from reading the PS file to determine
   % the bounding box (i.e., it will speed up the compilation process)
   % \includegraphics[width=.95\linewidth, bb=39 696 126 756]{sampleFig}
   %
  % \parbox[t]{.9\columnwidth}{\relax} deng adding some illustration.
   %
   \caption{ \label{fig:part_intepo_results}Smooth {interpolation} %interpolating 
   of the leftmost and rightmost samples with our implicit representations. }
 \end{figure}

{Previous works \cite{wu2019transgaga,chen2020deepfacedrawing} have shown that interpolation tends to give good beautification results. Furthermore, due to the continuous nature of our defined implicit function (hit function) for sketches over 2D space and sufficient training samples, our
%well-trained
{implicit sketch representation} %representations
work{s} well in producing smoothly interpolated and good results in the learned implicit field/space by linear interpolation.}
With the part-level {implicit sketch representations} of our dataset $\{v_1, v_2,\ldots,v_N\}$, we can synthesize the novel samples 
continuously and smoothly by linear interpolation {of implicit representations of existing sketches}, 
as shown in Figure \ref{fig:part_intepo_results}.
To further bridge the gap between the user's conceptual freehand sketches and the existing part sketches in our dataset, we adopt the same retrieval-and-interpolation strategy as \cite{chen2020deepfacedrawing} to instantiate the user's conceptual sketches with 
on-hand 
samples in our dataset by %{locally}
local 
linear interpolation.
{Specifically,} we first take the implicit representation $v_q$ of {an} %a 
input part sketch as the query to retrieve its top $K$ neighbors in our dataset. 
In the following steps, we construct a part-level manifold with the retrieved $K$ neighbors as {the} basis vectors $\{v_{t1},v_{t2},\ldots ,v_{tK}\}$, {and} project the $v_q$ to this manifold as Equation \ref{eq:manifold_projection}. 
\begin{equation}
   \label{eq:manifold_projection}
   \min{\lVert v_q - {\sum}_{i=1}^{K} w_i \cdot v_{ti}\rVert}^2, ~ ~ ~ ~ ~ ~ s.t. {\sum}_{i=1}^{K} w_i =1
\end{equation}
where we set $K = 3$ in our implementation and calculate the unknown weight parameter $w_i$ for each basis vector $v_{ti}$ by solving this constrained least-squares problem. {We finally
%, and 
obtain the projected implicit representation $v_p$ by {linearly} interpolating %linear-interpolating
the basis vectors locally: %, as Equation \ref{eq:local_linear_interpolation}. 
}
\begin{equation}
   \label{eq:local_linear_interpolation}
   v_p={\sum}_{i=1}^{K} w_i \cdot v_{ti}.
\end{equation}
Obtaining the projected implicit representation $v_p$, we further interpret it back to {the} 2D image space as the beautified reference for the input conceptual sketch (see Figure \ref{fig:fig_pipeline_sketch_beautification} {(c)}). 

\subsubsection{{Part-level} Geometry %b
Beautification}
\label{subsubsection:geometry_beautification}
\begin{figure}[tb]
   \centering
   % the following command controls the width of the embedded PS file
   % (relative to the width of the current column)
   \includegraphics[width=\linewidth]{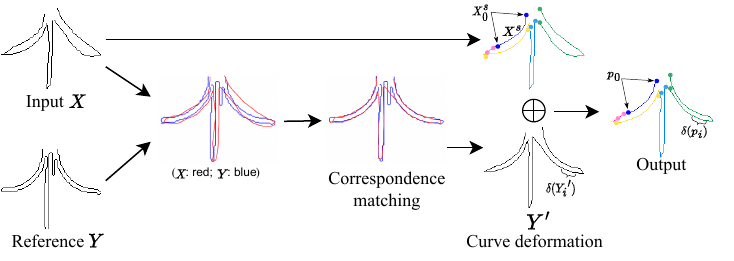}
   % replacing the above command with the one below will explicitly set
   % the bounding box of the PS figure to the rectangle (xl,yl),(xh,yh).
   % It will also prevent LaTeX from reading the PS file to determine
   % the bounding box (i.e., it will speed up the compilation process)
   % \includegraphics[width=.95\linewidth, bb=39 696 126 756]{sampleFig}
   %
  % \parbox[t]{.9\columnwidth}{\relax} deng adding some illustration.
   %
   \caption{\label{fig:geometry_beautification}
            {Illustration of our} pipeline for {part-level} geometry beautification. {Some results of Equations \ref{eq:2d_dynamic_registration} and \ref{eq:curve_deformation} are annotated here.}   
            }
 \end{figure}

In this work, we consider the {part-level} beautification as a deformation process that {deforms each part sketch} %interpolates the input 
towards {the corresponding} reference {obtained from the retrieval-and-interpolation step} rather than directly replacing the input with the reference{. In this way,} {%because
we hope to preserve the users' original drawing intentions as much as possible}. Thereby, given each input part sketch and its reference, we design the beautification pipeline as follow{s}: correspondence matching and curve deformation{. The former step performs a coarse shape-level registration that transforms all points of the input sketch to approximate the general shape of the reference.
Based on the output of correspondence matching, the {latter} %later
step then conducts a fine-grained curve-level deformation that deforms the input sketch towards {the} above intermediate output while {keeping} %keeps 
the original endpoints of the input curves unchanged}{, as illustrated in} %see
Figure
\ref{fig:geometry_beautification}.

For correspondence matching between the input part sketch and the reference sketch, we adopt a classical non-rigid registration method \cite{bouaziz2014dynamic}. 
Despite its robustness, this method fails to produce 
reasonable correspondence {results} efficiently when 
% being 
applied to our scenario.
We speed up this method 
{from more than 3 seconds} to around 0.3 seconds by adding two weight decay parameters $\alpha$ and $\beta$ that decrease exponentially with the increase of the iteration {$\eta$} 
as {shown in} Equation \ref{eq:2d_dynamic_registration}. Here we refer to the sketched points of {the} input and the reference as $X$ and $Y${, respectively}. 
\begin{equation}
   \label{eq:2d_dynamic_registration}
   \begin{aligned}
   & E_c=w_1E_{match} + {\alpha}^{\eta} \cdot w_2 E_{rigid} + {\beta}^{\eta} \cdot w_3 E_{arap},\\
   & E_{match} = {\sum}_{i=1}^{n}{\lVert z_i - P_y(z_i)\rVert}^2_2,               \\
   & E_{rigid} = {\sum}_{i=1}^{n}{\lVert z_i - R\cdot(x_i + t)\rVert}^2_2,     \\
   & E_{arap} =  {\sum}_{i=1}^{n} {\sum}_{j \in N_i} {\lVert (z_j-z_i) - R\cdot(x_j -x_i)\rVert}^2_2,
   \end{aligned}
\end{equation}
where $P_y(\cdot)$, $R$, and $t$ refer to the closest point in reference point set $Y$, the rotation matrix, and translate parameters,  respectively. {We initialize} 
$z_i$ %is initialized 
with $x_i$ in the input point set $X$, and %we 
set 
$w_1 =1, w_2=100,w_3 = 0.9$, $\alpha = 0.4$, and $\beta =0.9$. 
{Note that we solve the overall objective function $E_c$ in an iterative way.} In our experiment, 
{$\eta=$} 15 iterations are sufficient to produce a satisfying result {(see an example of $Y'$ in Figure \ref{fig:geometry_beautification})}.

To keep the user's original intention as much as possible, after calculating the 
{intermediate output {$Y'$} of the first step,} 
we further perform the curve deformation {that deforms} the recorded strokes 
% $S$
{$\{X^s\}$ ($s$ is the stroke index)} of the user's input {$X$} towards {$Y'$} 
{following} %as 
Equation \ref{eq:curve_deformation}. As we mentioned before, our interface can also record the strokes during {the} drawing process. 
\begin{equation}
   \label{eq:curve_deformation}
   \begin{aligned}
   &E_d =E_{position} + E_{shape}, \\
   &E_{position}=  {\sum}_{s \in S}
%   {\sum}_{i=1}^{n} 
   {\lVert p_0 -X_0^s\rVert},\\
   &E_{shape}= {\sum}_{s \in S}{\sum}_{i=1}^{n} {\lVert \delta(p_i) -\delta({Y_i}')\rVert}.  
   \end{aligned}
\end{equation}
Here, $\delta(k)=k_{i-1}+ k_{i+1} -2k_i$, {representing the local feature at each point}. 
{$X_0^s$ represents the {two} endpoints of a stroke 
% $s$
{$X^s$} of the input sketch $X$ while $p_0$ refers to the {corresponding} endpoints of the optimized curve. 
${Y_i}'$ represents the {closest}
point of 
% the intermediate output
$Y'$ to the point $p_i$ of the optimized curve under the Euclidean measurement {(see Figure \ref{fig:geometry_beautification})}.} 
We optimize the energy function in the stroke level.  The term $E_{position}$ enforces the optimized curve to have the same start point and the end point as {the} %user's 
input stroke, and the term $E_{shape}$ {constrains} %constraints
the optimized curve to have the same shape (curvature) as the corresponding reference stroke, as illustrated in Figure \ref{fig:geometry_beautification}. In this way, the final beautified sketch keeps the same start points and the end points {as those in the input sketch}.
We can think of our system as redrawing the {input} %user's 
strokes based on {their} %its 
original endpoints {in a more professional way}.

\begin{figure}[tb]
   \centering
   % the following command controls the width of the embedded PS file
   % (relative to the width of the current column)
   \includegraphics[width=.96\linewidth]{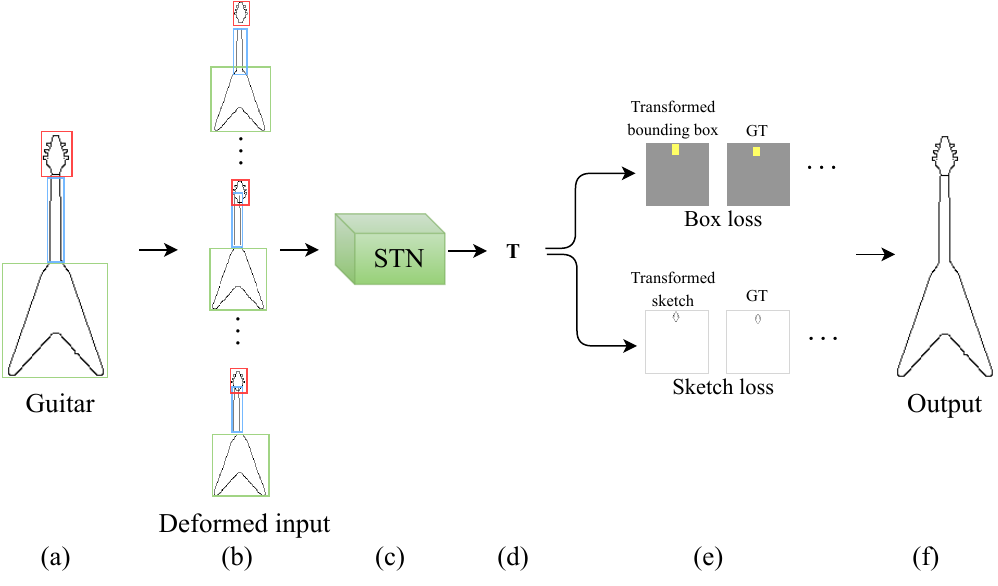}
   \caption{\label{fig:fig_part_assembly_pipeline}
           Illustration of our pipeline for learning  
           {structure beautification of the part-deformed} sketch.
           {(a) Ground truth. (b) Synthesized part-deformed sketches. (c) STN backbone of sketch assembly model. (d) Learned transformation matrix. (e) Losses for part-level bounding boxes and sketches. (f) Reassembled output.}}
 \end{figure}
% \subsection{Part Assembly}
\subsection{{Structure Beautification}}
\label{subsection: part_assembly}
Through the part beautification module, the geometry of input individual part sketches can be well refined. 
However, there is still a structure inconsistency existing in {the} %user's 
input sketch. 
To address this issue, {we further introduce a structure beautification module to adjust the imperfect structure of the input sketch with {the} deformed parts. The basis of {this} % the structure beautification 
module is the sketch assembly model designed to learn spatial transformations for beautifying the deformed sketches.}
{Figure \ref{fig:fig_part_assembly_pipeline} illustrates} %We show 
{the workflow of training}
our 
% part
{sketch}
assembly 
{model, where}
we first simulate the structure inconsistency by {randomly applying affine transformations to} %affining 
the {individual} part sketches in our dataset.
{In our implementation, we combine the affine transformation operations of random scaling (ranging from $0.8$ to $1.2$ folds) and random translations (varying in $x$ and $y$ directions with $-3$ to $3$ pixels offsets) in the image space {(see Figure \ref{fig:fig_part_assembly_pipeline} (b))}.}  
{As there is no obvious rotation observed in part sketches 
and our part beautification module is able to {reduce} %eliminate 
such rotation effects on part sketches, 
we do not apply the random rotations to our training data here.}
We then design a 
{sketch} % part
assembly network {(Figure \ref{fig:fig_part_assembly_pipeline} (c)) } with the backbone of the spatial transform network (STN, in short) \cite{jaderberg2015spatial} to learn the inverse mapping that transforms the distorted parts back to {their} %its 
ground-truth location{s} {(see Figure \ref{fig:fig_part_assembly_pipeline} (a))}. 
To prevent training 
% vibration
{oscillation}
caused by the sparse input (sketch), we further introduce the part-level bounding box{es} $M$ {(Figure \ref{fig:fig_part_assembly_pipeline} (e))} to enhance the spatial features of the part sketches $S$. 

\textit{Loss function.}
We
%deploy
use the {following} loss function for {training} the assembly network: % as follow\hb{s}:
\begin{equation}
   \label{eq:sk_assembly_loss}
   \begin{split}
   L= &{\sum}_{p \in P}~\lambda_1{\lVert S_p^{a} - S_p^{GT} \rVert} + \lambda_2{\lVert M_p^{a} - M_p^{GT} \rVert} \\  & \qquad \quad +  \lambda_3{\lVert T_p - T_p^{Identity} \rVert}^2_2,  
   \end{split}
\end{equation}
where we regard the three loss terms as sketch loss, bounding box loss, and {regularization} 
loss  
respectively. 
$S_p^a$ and $M_p^a$ are the part sketch and the part bounding box with the random affine transformation, respectively, %affine, 
while $S_p^{GT}$ and $M_p^{GT}$ are the {corresponding} ground truth in our dataset {(see Figure \ref{fig:fig_part_assembly_pipeline} (e))}. 
To stabilize the learning process, we further {constrain} %constraint 
a regularization {loss}
on the learned transformation matrix $T_p$ to enforce it to have a slow and mild update rather than a large 
{oscillation}
during the learning process.
$T_p^{Identity}$ refers to the Identity mapping matrix.
We set $\lambda_1=100$, $\lambda_2=1$, and $\lambda_3=1$ in the experiment.

%% file: sections/experiments.tex
\begin{figure}[tb]
   \centering
   % the following command controls the width of the embedded PS file
   % (relative to the width of the current column)
   \includegraphics[width=1\linewidth]{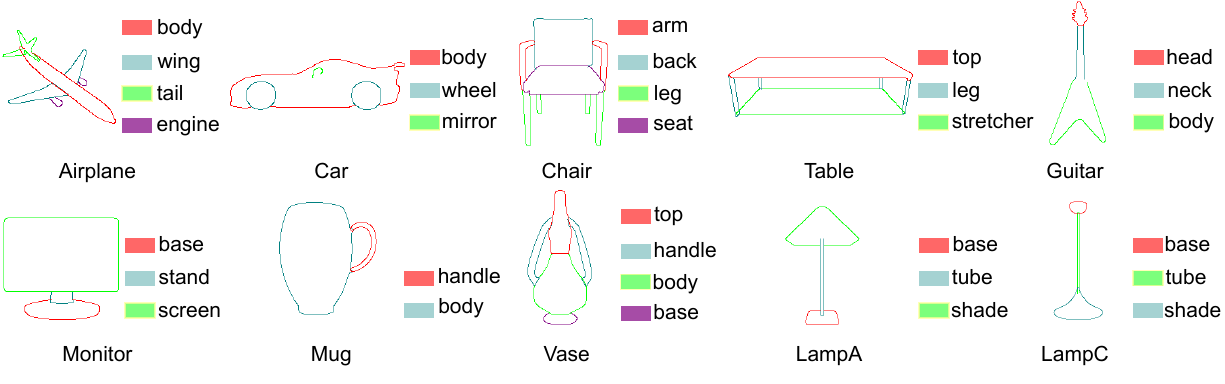}
   % replacing the above command with the one below will explicitly set
   % the bounding box of the PS figure to the rectangle (xl,yl),(xh,yh).
   % It will also prevent LaTeX from reading the PS file to determine
   % the bounding box (i.e., it will speed up the compilation process)
   % \includegraphics[width=.95\linewidth, bb=39 696 126 756]{sampleFig}
   %
  % \parbox[t]{.9\columnwidth}{\relax} deng adding some illustration.
   %
   \caption{\label{fig:dataset_images}
            Different categories and part annotations in our dataset. {Note that the lamp class has two different labeling strategies (see the part annotations of LampA and LampC).}}
 \end{figure}
\begin{table}
    \begin{center}
        \resizebox{\linewidth}{!}{
     \begin{tabular}{l c cc cc c c }
     \hline
     Categories & & &\#Samples & &\ \#Parts & & Source \\
     \hline
    \ \ Chair    & & & 5,962 & & \ 4 & & \ PartNet \ \  \\
    \ \ Table    & & & 4,440 & & \ 3 & & \ PartNet \ \ \\
    \ \ Airplane & & & 2,467 & & \ 4 & & \ \ \ \ SDM-Net \ \ \\
    \ \ Car & & & 1,813 & & \ 3 & & \ \ \ \ SDM-Net \ \ \\
    \ \ Guitar & & & \ \ \ 741   & & \ 3 & & \ \ \ \ SDM-Net \ \ \\
    \ \ Monitor & & & \ \ \ 559   & & \ 3 & & \ \ \ \ SDM-Net \ \ \\
    \ \ Vase     & & & \ \ \ 298   & & \ 4 & & \  COSEG \ \ \\
    \ \ Mug      & & & \ \ \ 211   & & \ 2 & & \  COSEG \ \ \\
    \ \ LampA    & & & \ \ \ 510   & & \ 3 & & \  COSEG \ \ \\
    \ \ LampC    & & & \ \ \ 171   & & \ 3 & & \  COSEG \ \ \\
     \hline
     \end{tabular}
        }
     \end{center}
    \caption{Data distribution of our synthesize{d} sketch dataset.
    }
    \label{table:data distribution}
\end{table}

\section{Experiment}
We conducted extensive experiments on  
freehand sketches drawn by {8} volunteers with our sketch interface. 
{Two of them were professional interior designers with years of drawing experience, and the rest were ordinary students aged 26 to 29 with no professional drawing skills. 
{Given an object category,} we asked the invited volunteers to sketch the man-made objects in their minds as casually as possible. 
They can draw any shape of part sketches around the common regions of the part shadows, like the airplane presented in the sketching canvas of Figure \ref{fig:fig_interface}.}
{Finally, we collected more than 200 freehand sketches and 15-45 sketched objects {that span diverse part geometry and global structures} for each category.}
Figures \ref{fig:teaser} and \ref{fig:fig_qualitative_comparison}
%for 
show
some
%{more}
representative drawings and the corresponding beautification results. {{Please find} more details of our collected freehand sketches and {additional} %the 
beautification results %are 
in the supplemental materials}. 
{We {also} construct a large dataset of part-labeled synthesized sketches to train the part-level sketch implicit representations and the sketch assembly model in our sketch beautification framework.} We show some rendered part-annotated sketches under the best view from the existing 3D shape repositories{,} {including} 
PartNet \cite{mo2019partnet}, SDM-Net \cite{gao2019sdm}, and COSEG dataset \cite{sidi2011unsupervised} {in Figure \ref{fig:dataset_images}}. % in 
Our sketch dataset contains 17,172 sketches with clear part annotations distributed over 9 man-made object categories {that are commonly observed in daily life} {\cite{eitz2012humans}}.
Figure \ref{fig:dataset_images} {gives a representative sketch under each category}.
We provide more details of the data distribution of our synthesized sketch dataset in Table \ref{table:data distribution}.

\subsection{Implementation Details}
We implemented our sketch implicit model and 
{sketch} % part
assembly model with the PyTorch framework \cite{paszke2019pytorch} and used the Xavier initialization \cite{glorot2010understanding}.
To train sketch implicit representations for our $128 \times 128$ part sketch, we sampled both the valid sketched points and the 
{invalid} % non-valid
points from the 2D image space
%in
(total 16,384 points).
We demonstrate the parameter structures of {the} sketch implicit model and 
% part 
{sketch}
assembly model in the supplemental materials. 
{The} two models {were} %are 
%both 
trained on an NVIDIA RTX 2080Ti GPU and optimized by the Adam  optimizer ($\beta$1 = 0.9 and $\beta$2 = 0.999) with the learning rate of $5e^{-5}$ and $1e^{-4}$, respectively.
Here we trained {the} two models to full convergence until the learning rate decayed to relatively small.
Note that training the sketch implicit model takes around 48 hours on a single GPU with the batch size of 1 for one category on average. The batch size for training 
% part
{sketch}
assembly is 64. 
The iteration epochs %in 
for the two models were %are 
set as 800 and 600{, respectively}. Although it takes a long time to train these two models during the training stage, our whole sketch beautification pipeline only spends around 1 second to beautify {an} 
entire sketch.

\subsection{Baselines}
\label{subsection: baselines}
To verify the effectiveness of our proposed sketch beautification pipeline, we compare our approach with existing methods both qualitatively and quantitatively. We use {four} baselines in our experiment and detail them as follows.

One of the baselines {is the Laplacian smoothing algorithm \cite{sorkine2004laplacian}. We implement this method on the stroke level to process the input freehand sketches. The next one}
is a naive instance-level and retrieval-based approach. 
We take the user's freely sketched input as the query and retrieve its top-1 candidate from the dataset as the beautified result. {In our implementation, we use the HOG \cite{dalal2005histograms} descriptor, {which is efficient and able to capture the spatial features of the sparse sketches.}}
Then, we further designed a part-level retrieval baseline.
{Different from instance retrieval using the entire input sketch as the query}, we first parse the user's freehand sketch to individual parts and utilize the separated part sketches (not the entire sketch) to retrieve the corresponding top-$1$ part results from the part-annotated dataset with the HOG descriptor, and finally replace the original part sketches with the retrieved part candidates as the final beautified output. 
Another baseline is a learning-based sketch simplification approach Mastering Sketching \cite{simo2018mastering}. 
{Although this generative method is designed to remove the superfluous details (i.e., the repetitive scribbles) and synthesize (or strengthen) the important lines for the rough sketches, we utilize it in our sketch beautification task to generate the beautified results for the input freehand sketches conditioned on the strokes that are identified as important by its learned model.}
In our implementation, we directly utilize their original trained inference model to process the user's freehand drawings. 
{Note that {since} the outputs of Mastering sketching are generated with bold effects, we %further 
post-process them with a thinning operation \cite{zhang1984fast} to remove the redundant pixels and conserve the clear skeletons as final beautification results.}
For the closely related work ShipShape \cite{fivser2016advanced}, since it was fully integrated into a quite old version of Adobe Illustrator and its plugin is no longer accessible today, we {can only discuss the difference between our approach and ShipShape (Section \ref{subsection: Sketch Beautification}) instead of conducting experimental comparisons}. %  cannot conduct an intuitive experimental comparison between ShipShape and our approach but only the more theoretical analysis in Section \ref{subsection: Sketch Beautification}.

\begin{figure*}[tb]
    \centering
    % the following command controls the width of the embedded PS file
    % (relative to the width of the current column)
    \includegraphics[width=\linewidth]{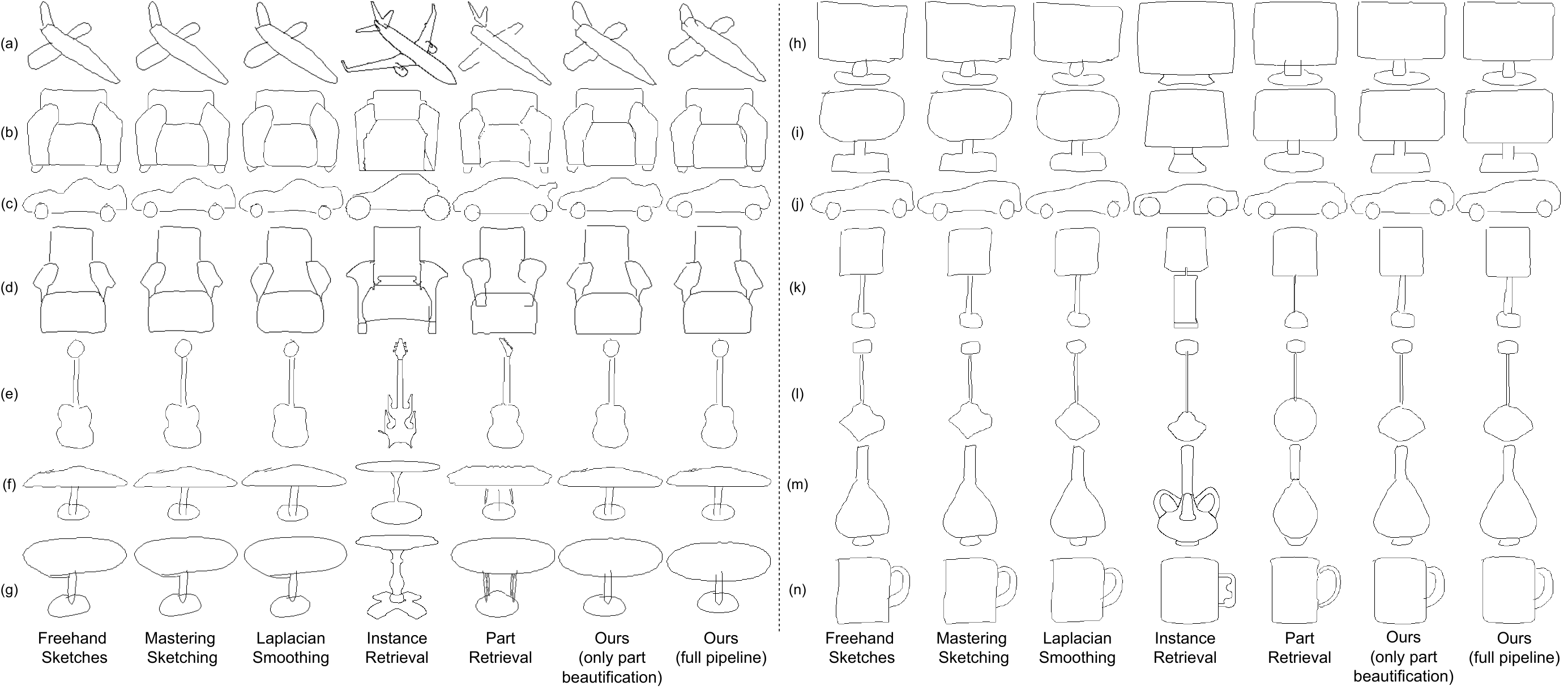}
    % replacing the above command with the one below will explicitly set
    % the bounding box of the PS figure to the rectangle (xl,yl),(xh,yh).
    % It will also prevent LaTeX from reading the PS file to determine
    % the bounding box (i.e., it will speed up the compilation process)
    % \includegraphics[width=.95\linewidth, bb=39 696 126 756]{sampleFig}
    %
   % \parbox[t]{.9\columnwidth}{\relax} deng adding some illustration.
    %
    \caption{\label{fig:fig_qualitative_comparison}            Comparison of different approaches {for} %in the task of 
    sketch beautification. Please zoom in for better visualization. }
  \end{figure*}
  
\subsection{Performance Evaluation}
We evaluate our method and compare it with the existing methods: {Laplacian smoothing,} instance retrieval, part retrieval, and Mastering Sketching 
% \cite{simo2018mastering} 
on freely drawn man-made object sketches.
{Figure \ref{fig:fig_qualitative_comparison} shows} 
% the
beautified results of different methods on the freehand sketches. {The visual comparisons show that} 
%Compared with other sketch beautification methods, 
our method produces the most satisfying results across various categories of sketches, regardless of the local geometry or the global structure.

We observe that the outputs of {Laplacian smoothing and }Mastering Sketching basically {keep} %keeping 
the same {shape} as the input sketches. {%However, 
Compared with the input, the former results lose some sharp features (see car body, chair seat, and guitar body in Figure \ref{fig:fig_qualitative_comparison} (c), (d), and (e) respectively) and the latter are generated with bold or shadow effects.}
Although Mastering Sketching succeeds in healing some small seams between parts ({e.g.,} %i.e., 
the head and neck parts of the guitar sketch in Figure \ref{fig:fig_qualitative_comparison} (e)) by 
% producing bold effects 
{generating more pixels (along the two strokes of the guitar neck towards the guitar head based}
on the original sketches), it fails to beautify the 
% more significant ones
{larger seams,} 
{e.g., defects between the neck and body parts of the 
guitar sketch in Figure \ref{fig:fig_qualitative_comparison} (e)}.

When visually inspected, the results produced by instance retrieval have the most dissimilar appearance compared with the input sketches.
Due to the freedom and abstraction of hand drawings {as well as the limited scale of our synthesized sketch dataset for retrieval}, instance retrieval cannot find a pleasing beautified counterpart from the existing sketch dataset for the user's input sketch.
Although part retrieval tries further to approximate the input sketches from a more fine-grained level, it only alleviates the dissimilar phenomenon. Still, these retrieval-based methods have no way to maintain the user's original drawing intentions.

To better demonstrate our sketch beautification method, we further present the intermediate outputs of the part beautification module in Figure \ref{fig:fig_qualitative_comparison}.
Unlike previous learning-based or retrieval-based methods,
our part beautification module produces {orderly lines, curves, and other primitives in}
part sketches
{by redrawing the input part strokes, leverages the knowledge of a large number of synthesized part-level sketches,} 
and also retains 
shapes that resemble
the original inputs without introducing extra strokes or missing any strokes. 
However, as shown in Figure \ref{fig:fig_qualitative_comparison}, some structure defects like seams (b), penetration (j), and misalignment (g) between part sketches still exist in the intermediate outputs of our part beautification module. 
This further emphasizes the importance of structural adjustment. When combined with the other component, our structure beautification module, {we} %the users
can obtain the most beautified outputs with pleasing part geometries and convincing global structures. 
{However, since there is no existing standard criteria to evaluate our proposed sketch beautification approach, 
we give our own insights and present the evaluation strategy 
as follows:
In fact, beautifying a sketch means adding changes to it while respecting the original sketch.
There is no perfect solution here,}
{{since} 
two main aspects ({i.e.,} improving beautification quality while respecting the original sketch) need to be balanced {during the beautification}.
{Regarding faithfulness to the input sketch and beautification quality,}
we further conduct a quantitative evaluation on the beautification faithfulness and a 
{perceptual} % perceptive
study on the beautification quality in %following 
Subsections \ref{subsubsection:faithfulness evaluation} and \ref{subsubsection: perceptive study}.
}

\begin{table}
    \begin{center}
     \resizebox{\linewidth}{!}{
     \begin{tabular}{l c c c c c c c c}
     \hline
     Methods   & & &mCD $\downarrow$  & & &mEMD($\times$ $10^{2}$) $\downarrow$ \\
     \hline
     Instance Retrieval  & & &15.55  & & & 5.92 \\
     Part Retrieval      & & &8.22   & & & 5.20 \\
     {Laplacian Smoothing} & & & {{3.44}} & & & {4.52}\\
      Mastering Sketching & & &\textbf{1.40} & & & \textbf{2.04}  \\
     Ours (only part {beautification}) % beauty) 
     & & & {4.59} & & &{4.62}     \\
     Ours (full pipeline)    & & & {6.84}  & & & {5.04} \\
     \hline
     \end{tabular}
    }
     \end{center}
    \caption{Quantitative evaluation on the faithfulness of the beautified results (produced by different sketch beautification methods) to the input sketches.}
    \label{table:table_quantitative_evaluation}
\end{table}

\subsubsection{{Quantitative Evaluation on Faithfulness}}
\label{subsubsection:faithfulness evaluation}

To {quantitatively} %further 
evaluate the performance of different {beautification} approaches {in preserving the user's original drawing intentions} (i.e., faithfulness), 
we report the statistic values of two metrics for the aforementioned methods in Table \ref{table:table_quantitative_evaluation}.
{To measure the difference between a beautified result and the user's original freehand sketch,}
we adopt the Chamfer Distance-L2 (CD) and Earth Mover's Distance (EMD) as the evaluation metrics (lower is better) for the {faithfulness evaluation in} the sketch beautification task. {The former metric is employed to measure the point-wise distance between the sketched objects in two sketches and the latter one is utilized to compute the distribution-level distance of two point distributions over the entire image space ($256 \times 256$)}.  
Given a pair of the user's freely sketched input and the beautified output produced by {one of the compared methods}, 
we calculate the Chamfer Distance of the only valid {sketch} % sketched 
pixels ({i.e., excluding} %not including 
the background pixels) in {the} two sketches. 
When computing the Earth Mover's Distance, we preserve both the {sketch}  
pixels and the background pixels and treat the pair of {the} input and output sketches as two distributions{.}
Finally, we average these two metrics over pairs {of {sketches before and after beautification} {in our collected freehand sketch dataset}} and present the mean values of CD and EMD in Table \ref{table:table_quantitative_evaluation}.

Consistent with what is observed from Figure \ref{fig:fig_qualitative_comparison},
{as shown %demonstrated 
in Table \ref{table:table_quantitative_evaluation},}
our method outperforms its retrieval-based competitors while falling behind {Mastering Sketching and} Laplacian smoothing quantitatively. 
{It is reasonable that the results of {Mastering Sketching}
 have the closest distance to the input sketches since this method {inherits (or accepts) all {the input} strokes 
 with limited pixel-level beautification.}}
For {Laplacian smoothing}, since this method only slightly changes the {position} of the sketched points by smoothing the local-level strokes{, it is easier for this method to achieve a remarkable performance (the second {ranking}) in faithfulness evaluation {(see Table  \ref{table:table_quantitative_evaluation})}.} 
While the fine-grained part retrieval {outperforms} %can boost 
the instance retrieval significantly, 
{it is still inferior to} %are still beneath 
our method.   
The beautified outputs of our part beautification module and our full pipeline achieve competitive performance to  Laplacian smoothing in preserving 
% user's
{the} user's
original drawing intentions
(the {third} and the {fourth} {rankings} in Table 2). 
It is expected that the results of our part beautification module have the closer distance to the user’s input sketches compared with our full pipeline {(as reported in Table \ref{table:table_quantitative_evaluation})} since the part beautification module only beautifies the curve shape of the part sketches without adjusting the scales or locations of the part sketches. 
Just as we discussed before, there are still a lot of structure errors ({e.g.,  seams, penetration, and misalignment)} in the intermediate beautified outputs, as shown in Figure \ref{fig:fig_qualitative_comparison} (Ours (only part beautification)).
{To correct such structural errors {during the structure beautification stage}, a small range of structural refinement (including pixel-wise translations and scalings) is applied to the beautified part sketches. % during the structure beautification stage. 
Therefore, it is inevitable that our full pipeline combining the part beautification and structure beautification modules slightly enlarges the distance to the {input} %user's
freehand sketches.
For the faithfulness aspect of the sketch beautification task, since our framework is proposed to beautify both the local geometry and global structure of input sketches, our method cannot keep the input sketches almost unchanged as in Mastering Sketching.
But our approach does not heavily change the input sketches as instance and part retrieval, and can also be regarded as a larger 
``{enhancement}'' operation 
({making the original input sketches  aesthetically more beautiful} in both local curve and global shape) to some extent.} 
For the evaluation on the beautification quality of the different sketch beautification methods, we demonstrate it with a user study in the next section.

\begin{figure}[tb]
    \centering
    % the following command controls the width of the embedded PS file
    % (relative to the width of the current column)
    \includegraphics[width=0.7\linewidth]{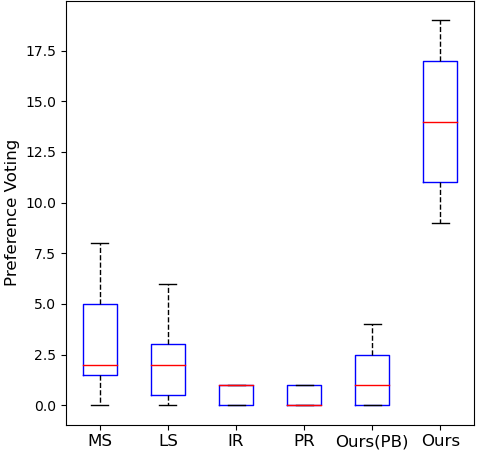}
    % replacing the above command with the one below will explicitly set
    % the bounding box of the PS figure to the rectangle (xl,yl),(xh,yh).
    % It will also prevent LaTeX from reading the PS file to determine
    % the bounding box (i.e., it will speed up the compilation process)
    % \includegraphics[width=.95\linewidth, bb=39 696 126 756]{sampleFig}
    %
   % \parbox[t]{.9\columnwidth}{\relax} deng adding some illustration.
    %
    \caption{\label{fig:fig_user_study} 
    % (a) 
    Box plots of the average preference voting over the prepared questions (beautification quality) for each method. 
    % {MS, LS, IR, PR, Ours (PB) and Ours} are the six compared approaches.
    {
    % The six compared approaches in this box plot, namely, 
    MS, LS, IR, {and} PR stand for Mastering Sketching, Laplacian smoothing, Instance retrieval, {and} Part retrieval{, respectively}.
    %And 
    Ours (PB) and Ours refer to the part beautification module and the full pipeline of our sketch beautification approach, respectively.}}
  \end{figure}  
  
\subsubsection{{Perceptual}
% Perceptive
Study {on Beautification {Quality}}}
\label{subsubsection: perceptive study} 
It is known that the concept of beautification is highly relevant to human preference and perception.
Therefore, to evaluate the beautification quality of the beautified sketches produced by different methods in Section \ref{subsection: baselines} (namely, Laplacian smoothing (LS), Mastering Sketching (MS), instance retrieval (IR), part retrieval (PR), {our part beautification module only (Ours (PB))} and our {full} approach), 
we further conducted a user study to evaluate the performance of these approaches in the sketch beautification task.

Specifically, we first randomly picked a set of 15 input sketches from the drawn 
freehand sketches from all the categories in our dataset. 
We then applied the aforementioned {six} sketch beautification methods to each input sketch to generate the beautified results.
Figure \ref{fig:fig_qualitative_comparison} displays some representative examples of inputs and outputs used in our user study. 
The evaluation was done through an online questionnaire. 
There were in total {23} participants (15 males and 8 females) in this study.  
We showed each participant sets of {an input sketch and} {six}
beautified sketches generated by the compared approaches in random order set by set. 
Each participant was asked to select the most beautiful result {in each set} regarding the visual aesthetics from both aspects of the local geometry and the global structure. 
Finally, we received {23} (participants) $\times$ 15 ({input} sketches) 
= {345} subjective evaluations.

Figure \ref{fig:fig_user_study} 
% (a) 
{shows} 
%demonstrates 
the statistics of the voting results. 
We conducted one-way ANOVA tests on the preference voting results, and found significant effects for aesthetic preference to our method ($F =  65.88$, $p < 0.001$).
Then further paired t-tests show that our method (mean: $13.93$) got significantly more votes than all the other methods, Mastering Sketching (mean: $3.13$, [$t = 2.14$, $p < 0.001$]), Laplacian smoothing (mean: $2.47$, [$t = 2.14$, $p < 0.001$]), instance retrieval (mean: $1.00$, [$t = 2.14$, $p < 0.001$]), part retrieval (mean: $0.47$, [$t = 2.14$, $p < 0.001$]), and our part beautification module (mean: $2.00$, [$t = 2.14$, $p < 0.001$]).

%To sum up,
To summarize,
%for the evaluation on the beautification quality of the compared methods,
our approach achieved the best performance that {significantly}
% largely
surpasses its competitors under human aesthetic perceptions.
Although there are {slightly higher} %some mild 
deviations in the beautified results generated by our method compared with the original input sketches, our beautified sketches are still able to be faithfully {voted as the best beautification results for the given sketches} by the users. In addition, our approach successfully beautifies the local-level part geometry and corrects the global-level structural errors of input sketches. % , which
{In this way, our proposed method} 
helps to improve the beautification quality of input %users' 
freehand sketches significantly. 
We show more beautification results of our method (including the intermediate outputs after part beautification and final results after structure beautification) in the supplemental materials.
%for the interested readers.
% During the data analysis stage, we found a notable case (Figure \ref{fig:fig_user_study} (b)) where our method received far fewer votes (2 votes) than Mastering Sketching (12 votes). 
% As shown in Figure \ref{fig:fig_user_study} (b), given the same input, our method beautifies the wheel part of the input sketch from the triangle shape towards the circle shape;
% Instance and part retrieval methods beautify the triangle
% wheel by directly replacing it with the retrieved circle wheel; 
% In the above methods, the original geometry of the sketched wheel is totally washed away. 
% On the contrary, Mastering Sketching keeps the triangle-shaped wheel but fails to beautify it.
% This indicates two purposes of beautifying input sketches and preserving the user's original drawing intention might be mutually contradictory and fail to achieve simultaneously when the user draws a very poor and strange input sketch.

% In this paper, we 

% This is the common limitation of learning-based methods, since the triangle-shape wheel 

% We further interviewed several participants after the study and asked why they voted for Mastering Sketching rather than our approach in this case.
% One participant said "Considering the geometry aspects, it is supposed to choose Mastering Sketching since the rest candidates do not look like the input sketch". 

\begin{table}
    \begin{center}
    \resizebox{\linewidth}{!}{
     \begin{tabular}{l l c c c}
     \hline
     Input &  Loss &mIOU(\%)$\uparrow$  &mCD$\downarrow$ & mEMD($\times$$10^{2}$)$\downarrow$ \\
    
     \hline
     sk&skloss   &0.00 & $+\infty$ & $+\infty$ \\
     sk&(sk+regu)loss  &83.38 &5.85 &6.75\\
     sk+bb&bbloss  &91.87 &2.79 &5.42\\
     sk+bb&(sk+bb)loss   &91.83 &2.79 & 5.42\\
     sk+bb&(sk+bb+regu)loss  &\textbf{91.89}&\textbf{2.78} &\textbf{5.40}\\
     \hline
     \end{tabular}
    }
     \end{center}
    \caption{Ablation study of our designed mechanism for the sketch assembly model. 
    The sk and bb in the first column are the inputs of the sketch and the bounding box, respectively. 
    The skloss, bbloss, and reguloss in the second column refer to the supervision losses of the sketch, the bounding box, and the regularization term{, respectively}. 
    }
    \label{table:tabablation_loss}
\end{table}

\subsubsection{Ablation Study} 
{Since} we have demonstrated the roles of the part beautification module and the structure beautification module of our sketch beautification pipeline qualitatively and quantitatively in {the} above two subsections, and detailed the key components of the part beautification module in Section \ref{subsection: part_beautification}{, we} 
do not conduct the redundant ablation studies for the part beautification module here.

Hence, in this subsection, we mainly focus on validating the effectiveness of the key components in the structure beautification module. 
Since the sketch is a kind of sparse representation, it is a nontrivial task for our sketch assembly model to represent and learn its spatial transformations precisely and effectively.
To
%qualify
validate our designed mechanism of the sketch assembly model, we perform ablation studies of its key components in turn, namely, the sketch loss, the bounding box loss, and their combinations with the regularization loss (see Table \ref{table:tabablation_loss}).

\begin{figure}[tb]
    \centering
    % the following command controls the width of the embedded PS file
    % (relative to the width of the current column)
    \includegraphics[width=1\linewidth]{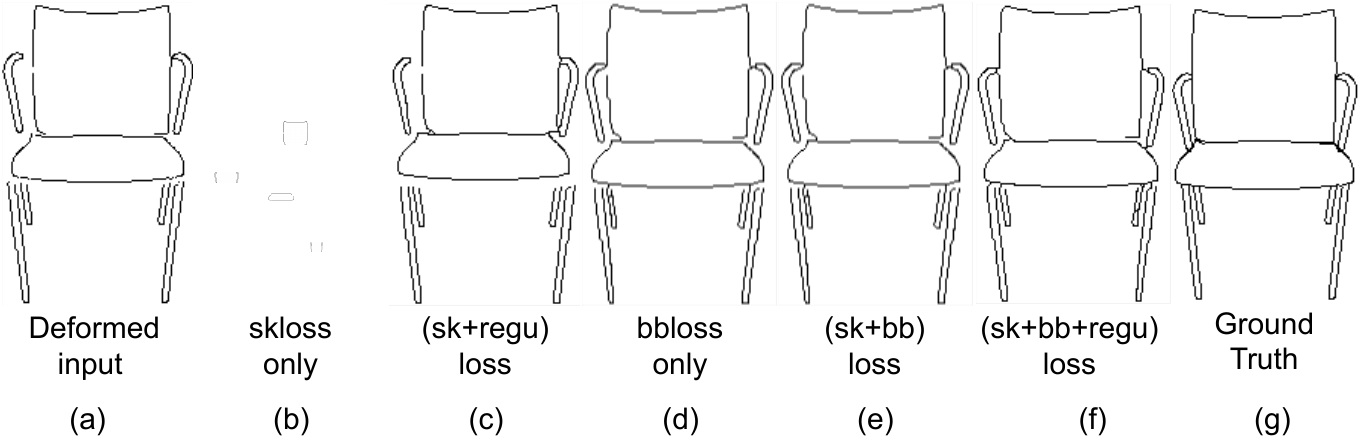}
    % replacing the above command with the one below will explicitly set
    % the bounding box of the PS figure to the rectangle (xl,yl),(xh,yh).
    % It will also prevent LaTeX from reading the PS file to determine
    % the bounding box (i.e., it will speed up the compilation process)
    % \includegraphics[width=.95\linewidth, bb=39 696 126 756]{sampleFig}
    %
   % \parbox[t]{.9\columnwidth}{\relax} deng adding some illustration.
    %
    \caption{\label{fig:fig_ablation_study}  Visual comparison of different supervision losses during the training process. Please zoom in for better visualization (in particular for ``skloss only'').}
  \end{figure}

Specifically, we use 1,000 randomly sampled synthesized sketches of the Chair category
as the ground truth since this category
%owns
has the most complex and challenging structures in our dataset.
We then apply random affine transformations to the parts of the ground truth samples five times
and take these 5,000 deformed sketches with the warped parts as the benchmark input. 
Finally, we evaluate the performance of the trained sketch assembly models under different training strategies by measuring the disparity between their assembly outputs and the ground truth.
{Note that although the 1,000 ground truth sketches are within the same set used for retrieval in our beautification pipeline in Section \ref{subsubsection: Retrieval_and_Interpolation}, the benchmark (5,000 deformed sketches) for testing here is already significantly different from the original set after five times of random affine transformations.}
To better reflect the performance of the methods in the sketch assembly task, we compute a region-based and part-level IOU metric on bounding boxes of the transformed part outputs and the corresponding part ground truth, as {formulated} 
%illustrated 
in {the following equation}: % Equation \ref{eq:IOU metric}: 
\begin{equation}
   \label{eq:IOU metric}
   IOU= \frac{1}{N_p}\cdot {\sum}_{p \in P}~ {\frac{M_p^a \cdot M_p^{GT}}{M_p^a + M_p^{GT}}}, 
\end{equation}
where $N_p$ is the {number} % amount 
of %the 
parts $P$ in a sketch, $M_p^a$ and $M_p^{GT}$ are the part-level bounding boxes of the assembled sketches and ground-truth sketches, respectively.
We also calculate the Chamfer Distance-L2 (CD) and Earth Mover’s
Distance (EMD) between the assembled and ground-truth sketches to further verify the performance of the ablated sketch assembly models.
Table \ref{table:tabablation_loss} {reports} %plots 
the mean values of the above metrics.
We also show the performance of different components in the sketch assembly task qualitatively in Figure \ref{fig:fig_ablation_study}.

% \subsection{Perceptive Study}
% To be written

In our experiments, we found that only utilizing the sketch loss cannot supervise the learning process of spatial transformations of sparse sketches. 
Due to limited valid pixels in sketches, the network tends to shift its focus from sketched pixels to the background pixels during the training process, even under the L1 loss of the input and output sketches. 
With {the number of} training epochs increasing, the model degrades rapidly and ignores the sketched pixels until the part sketches are rescaled to none. 
Hence, the mean IOU value of 'skloss only' model in Table \ref{table:tabablation_loss} is $0$. %And 
The other two metrics also show the failure of training on sketch input {with} %and 
the sketch loss {only}.
This can be further witnessed in Figure \ref{fig:fig_ablation_study} (b).
Note that we {screen-captured} %screenshoted 
this subfigure of ``skloss only'' at the very beginning of {training} %epochs 
since the part sketches {would} % will 
disappear soon after several further epochs. 
Along with the degradation of ``skloss only'' model, we observed a significant increment in the parameters of the learned transformation matrices. 
We further designed and added the regularization loss to penalize this huge 
% vibration
{oscillation}
of the learned transformation matrices. 
However, although the input part sketches would no longer shrink to nothing by {applying}
% impelling
the additional regularization loss {to}
% over
the learned transformation matrices, the network still failed to learn meaningful spatial transformations (just random translation or scaling under the constraint of the regularization loss), as shown in Figure \ref{fig:fig_ablation_study} (c).
Only by introducing the bounding box input and the corresponding bounding box loss, the networks were able to learn the spatial transformation of sparse sketches stably and effectively (see the inputs containing 'bb' and the losses with 'bbloss' in Table \ref{table:tabablation_loss} and Figure \ref{fig:fig_ablation_study} (d-f)). 
%As we combined 
With the combination of the sketch loss, the bounding box loss, and the regularization loss, the network achieved the best performance, as shown in Table \ref{table:tabablation_loss} quantitatively and Figure \ref{fig:fig_ablation_study} (f) qualitatively.
These results further confirm the necessity and importance of our design choices for the structure beautification module.

%% file: sections/conclusion.tex
\section{Conclusion {and Future Work}}
We have introduced {an} intuitive and {effective}
% and generic
beautification approach for freehand sketches depicting man-made objects by conducting {part-level} %local 
geometry beautification and global structure refinement {sequentially}. 
As one of the key components in our approach, the sketch implicit model can be easily plugged into contemporary deep neural networks for a variety of tasks relevant to sketches including sketch recognition, classification, retrieval, reconstruction, and generation thanks %owe 
to its promising representation capacity {in precisely reconstructing the input sketch (Figure 2), and in generating novel, smooth, and continuous samples by
interpolating among two/multiple existing sketches (Section 3.2.2).}
The other component, i.e., the sketch assembly model, provides a robust and effective solution for compositing sparse 2D components by
%adding
having the part-level bounding boxes.
The whole beautification pipeline could further inspire and boost the downstream applications that operate over %the 
freehand sketches as input.
However,
%to
beautifying a freehand sketch under {an} 
{arbitrary} view is still challenging for our method. 
{Furthermore, the voting results in our perceptual study (Figure 11) indicate that structure beautification has the strongest effect on user preference when perceiving beautification quality. 
Therefore, further exploration of how and which factors (e.g., more obvious spatial changes) cause such a preference difference when users perceive geometry and structure changes is a valuable and promising direction for beautification tasks.}
We leave {these} {for} future work to explore.

\begin{figure}[t]
   \centering
   % the following command controls the width of the embedded PS file
   % (relative to the width of the current column)
   \includegraphics[width=1\linewidth]{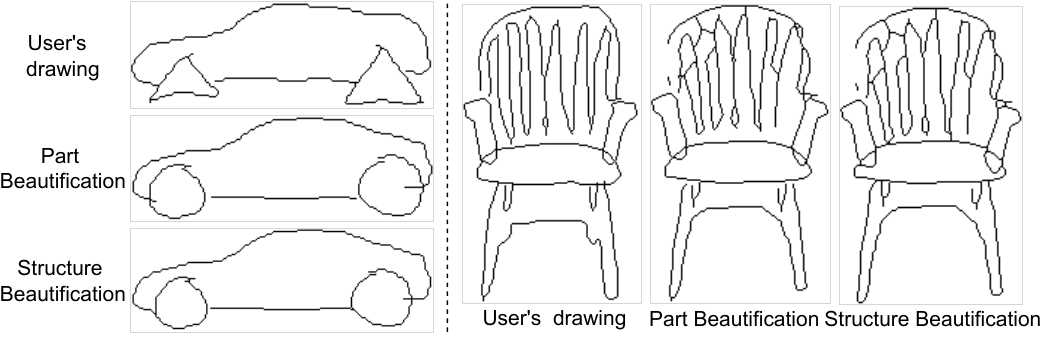}
   % replacing the above command with the one below will explicitly set
   % the bounding box of the PS figure to the rectangle (xl,yl),(xh,yh).
   % It will also prevent LaTeX from reading the PS file to determine
   % the bounding box (i.e., it will speed up the compilation process)
   % \includegraphics[width=.95\linewidth, bb=39 696 126 756]{sampleFig}
   %
  % \parbox[t]{.9\columnwidth}{\relax} deng adding some illustration.
   %
   \caption{\label{fig:fig_limiations}
           {Failure cases of our sketch beautification method.} 
           }
 \end{figure}

%% file: sections/limitation_and_discussion.tex
%\section{Limitations and Discussions}

%\hbc{Maybe you can move the discussions on the limitations and possible solutions after the conclusion}\dengc{updated}
%{Our method suffers from a few limitations.}
Our method has some limitations.
%In this section, we will discuss the limitations of our method and the potential solutions.
First, in our method, all the training samples of the sketches of man-made objects are rendered under the canonical view (best view) for each category.
This limits the adaptation ability of our method to freehand sketches with large view variations or multi-view sketches. 
Although this 
% could
{is}
% be 
%\hbc{`could' means you can do but you haven't done it yet. `is'?} \dengc{updated.} 
partially solved by adding a shadow guidance to 
% constraint
{constrain}
%\hbc{`constraint' is a noun. use `constrain'} 
the user's input, we {are interested in addressing} % still look to address 
this issue in a more elegant way \cite{yu2020sketchdesc}.
Second, a key merit of our method is the representative power of the sketch implicit representations that can interpolate the sketched points smoothly and continuously.
But this implicit representation {requires} %consumes 
a longer training time than the CNN representations on the same input data, since the implicit model needs to sample all the coordinates and remember the ground truth value of each point in the 2D image space. 
One possible solution {for speeding up the training process} %\hbc{for speeding up the training process?} 
is to change the sampling strategy, {e.g., by keeping} % that keep
sampling the points close to the strokes instead of sampling all the points from {the} whole 2D space.
% Lastly, %At last, 
{Third,} %\hbc{use `First', `Second', `Third'}
while our method is able to instantiate %the 
conceptual freehand sketches, 
%if %the 
%input sketches are drawn too bad{ly} or too complex{ly}, 
our approach could fail{ if input sketches are drawn too
%badly
poorly or {too complex{ly}}, as shown in Figure \ref{fig:fig_limiations}.
%\hbc{You meant two examples in this figure, or one of them?}}\dengc{'too complexly' was wrongly commented by me.
}
{If a user draws a very 
% poor
{unnatural}
sketch (see the triangle-like wheels of a car on the left-top,
%\hbc{In this case, it is very likely what the user wants is triangular wheels. I don't think this should be considered as a poor sketch? It is more related to the lack of similar examples in the training dataset?}\dengc{Yes. I changed the 'poor' to 'unnatural'. }), 
our method might not follow the user's drawing intention and even totally changes the geometry of part sketches drawn by the user. This violates the beautification constraints we set in this paper (as {discussed} %demonstrated 
in
%the fourth paragraph of
Section \ref{section:introduction}).
If a user sketches a very complex part (see the chair back with too many sticks on the right), our approach also cannot beautify such a %that hard 
part.}
This is due to the {failure of the} correspondence matching step in {Section} % the section
\ref{subsubsection:geometry_beautification}.
It is known that correspondence matching is still 
% a
{an}
%\hbc{again, a minor but common grammar problem} 
open problem in the research community. Therefore, the improvement on registration and correspondence matching could also further boost the performance of our method.
%{Lastly, as humans' perception and
%preference for beautification concepts are %similar but %are 
%not exactly the same (shown in
%Figure \ref{fig:fig_user_study} in our %\deng{perceptual} 
%study),
Lastly, users should ideally be allowed to adjust the degree of beautification (more significant beautification would lead to larger deviation from the input sketch).
In our approach, the beautification function is designed in a closed and automatic way for efficiency reasons.
In the future, we would like to extend our sketch beautification approach to allow for more user control.
%to the user-controlled version.
%}

\section*{Acknowledgements} 
We thank the anonymous reviewers for their constructive comments. This work was partially supported by grants from the Research Grants Council of the Hong Kong Special Administrative Region, China (No. CityU 11212119, 11206319, and 11205420), the Chow Sang Sang Group Research Fund/Donation, and the Centre for Applied Computing and Interactive Media (ACIM) of the School of Creative Media, CityU.